\pdfoutput=1

\documentclass[11pt]{article}

\usepackage[preprint]{acl}
\usepackage{times}
\usepackage{latexsym}
\usepackage[T1]{fontenc}
\usepackage[utf8]{inputenc}
\usepackage{microtype}
\usepackage{inconsolata}
\usepackage{graphicx}
\usepackage{booktabs}
\usepackage{multirow}
\usepackage{amsmath}
\usepackage{colortbl}
\usepackage{tabularx}
\usepackage{xcolor}
\usepackage{xspace}
\usepackage{cleveref}
\usepackage{adjustbox}
\usepackage{hyperref}

\newcommand{\gpt}{\textsc{GPT-4o-mini}\xspace}
\newcommand{\gpta}{\textsc{GPT-4o}\xspace}
\newcommand{\intern}{\textsc{InternVL2-26B}\xspace}
\newcommand{\data}{\textsc{Prism}\xspace}

\newcommand*\samethanks[1][\value{footnote}]{\footnotemark[#1]}
\newcommand*\corresponding[1][3]{\footnotemark[#1]}

\title{Vulnerability of Text-to-Image Models to Prompt Template Stealing: A Differential Evolution Approach}

\author{
 \textbf{Yurong Wu\textsuperscript{1,3}\thanks{Equal contribution.}},
 \textbf{Fangwen Mu\textsuperscript{1,3}\samethanks[1]},
 \textbf{Qiuhong Zhang\textsuperscript{1,3}\samethanks[1]},
 \textbf{Jinjing Zhao\textsuperscript{4}},
 \textbf{Xinrun Xu\textsuperscript{1,3}},
\\
 \textbf{Lingrui Mei\textsuperscript{3}},
 \textbf{Yang Wu\textsuperscript{3}},
 \textbf{Lin Shi \textsuperscript{1,3}},
 \textbf{Junjie Wang\textsuperscript{1,3}},
 \textbf{Zhiming Ding\textsuperscript{1,3}\thanks{Corresponding author.}},
 \textbf{Yiwei Wang\textsuperscript{2}\corresponding[2]}
\\
 \textsuperscript{1}Institute of Software, Chinese Academy of Sciences\quad
   \textsuperscript{2}University of California at Merced\\
   \textsuperscript{3}University of Chinese Academy of Sciences\quad
    \textsuperscript{4}The University of Sydney\\
 \{wuyurong20, zhangqiuhong22, xuxinrun20, meilingrui22\}@mails.ucas.ac.cn\\
 \{fangwen2020, shilin, junjie, zhiming\}@iscas.ac.cn\quad jzha0100@uni.sydney.edu.au\\
 wuyang2023@ia.ac.cn \quad wangyw.evan@gmail.com
}

\begin{document}
\maketitle
\begin{abstract}
Prompt trading has emerged as a significant intellectual property concern in recent years, where vendors entice users by showcasing sample images before selling prompt templates that can generate similar images. This work investigates a critical security vulnerability: attackers can steal prompt templates using only a limited number of sample images. 
To investigate this threat, we introduce \textbf{\data}, a prompt-stealing benchmark consisting of 50 templates and 450 images, organized into Easy and Hard difficulty levels.
To identify the vulnerabity of VLMs to prompt stealing, we propose \textbf{EvoStealer}, a novel template stealing method that operates without model fine-tuning by leveraging differential evolution algorithms. The system first initializes population sets using multimodal large language models (MLLMs) based on predefined patterns, then iteratively generates enhanced offspring through MLLMs. During evolution, EvoStealer identifies common features across offspring to derive generalized templates. 
Our comprehensive evaluation conducted across open-source (\intern) and closed-source models (\gpta and \gpt) demonstrates that EvoStealer's stolen templates can reproduce images highly similar to originals and effectively generalize to other subjects, significantly outperforming baseline methods with an average improvement of over 10\%. Moreover, our cost analysis reveals that EvoStealer achieves template stealing with negligible computational expenses. Our code and dataset are available at~\url{https://whitepagewu.github.io/evostealer-site}.
\end{abstract}

\section{Introduction}
\begin{figure}
    \centering
    \includegraphics[width=\linewidth]{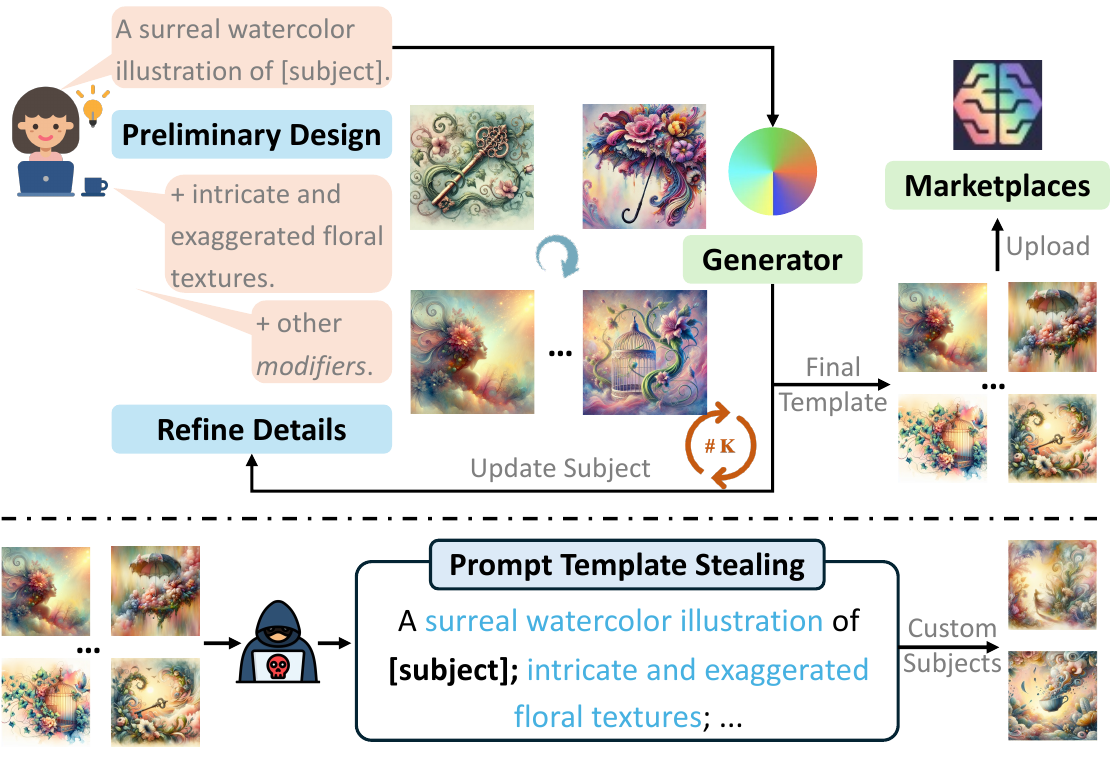}
    \caption{Top: Illustrating the legitimate development of text-to-image prompt templates. Bottom: Depicting unauthorized extraction of proprietary prompt templates}
    \label{fig1}
\end{figure}

Recent advancements in text-to-image generation~\citep{liu2024visual,cao2024controllable}, particularly in multimodal large language models (MLLMs)~\citep{liu2024visual,wang2024qwen2} and diffusion models~\citep{ho2020denoising,sohl2015deep}, have significantly improved image generation performance. 
However, crafting the perfect prompt to produce desired output images results remains a meticulous process that requires significant expertise and time investment~(Refer to Figure~\ref{fig1}(top))~\cite{wu2024strago,yang2024ampo}.
This challenge has catalyzed the emergence of prompt trading, a novel business model exemplified by platforms like PromptBase~\footnote{https://promptbase.com/} and LaPrompt~\footnote{https://laprompt.com/}. On these platforms, creators upload meticulously crafted prompt templates~(viewable post-purchase) alongside multiple sample images~(publicly visible). Customers attracted to these samples can purchase the template, then merely modify the subject specification to generate new images that preserve the original stylistic elements.
In this context, the platform’s copyright and security vulnerabilities raise significant concerns. If attackers reverse-engineer the proprietary templates by analyzing the visible samples, they could significantly compromise sellers' intellectual property rights and threaten the platform's business model~(Refer to Figure~\ref{fig1} (bottom)). We term this attack \textit{prompt template stealing}.

Existing methods for prompt stealing attacks~\citep{shen2024prompt,sha2024prompt,naseh2024iteratively} focus on reconstructing individual prompts for each sampled image, rather than recovering a general prompt template for the entire group of sampled images. As a result, the prompts reconstructed by these methods are specific to each image and lack generalizability, which limits their applicability in practical scenarios, as illustrated in Figure~\ref{fig1}. For example, in the case of the woman image located in Figure~\ref{fig1}, a stolen prompt might include the "golden sun" as a distinctive element. Nevertheless, a comparison with the other three images demonstrates that the "golden sun" is not a shared characteristic among them.

To fill this gap,  we build a new and comprehensive benchmark named \data, comprising 50 prompt templates stratified across two difficulty levels (Easy and Hard) and spanning 9 distinct subjects, sourced from a specialized prompt trading platform. 
Utilizing DALL·E 3, we generated 450 images, with each group methodically partitioned into 5 in-domain and 4 out-of-domain images to systematically evaluate both model fitting capability and generalization performance. 

Besides, we introduce EvoStealer, a novel template stealing methodology derived from the differential mutation algorithm in evolutionary computation. Our approach strategically leverages mutation and crossover operations within the search space to effectively mitigate overfitting and circumvent local optima, precisely aligning with template stealing objectives. We integrate large language models (LLMs) spanning both open-source and closed-source domains, specifically utilizing InternVL2-26B, GPT-4o, and GPT-4o-mini. By combining these models with a differential evolution algorithm, we generate prompt templates characterized by exceptional stability and robust generalization capabilities. Comprehensive experimental evaluations are conducted across easy and hard difficulty levels. The results demonstrate EvoStealer's remarkable performance: the methodology efficiently reproduces images highly similar to original templates while simultaneously exhibiting strong cross-subject generalizability. This enables large-scale image generation maintaining consistent stylistic characteristics.

Our main contributions are as follows:

(1) To the best of our knowledge, this is the first systematic study on prompt template stealing, revealing its severity as an emerging security threat and empirically demonstrating its significant risk to intellectual property protection;

(2) This study introduces \data, the first benchmark for prompt template stealing, and EvoStealer, a plug-and-play attack framework that requires no fine-tuning, significantly improving practicality and scalability;

(3) We conducted extensive experiments on both open-source models (\intern) and closed-source models~(\gpta, \gpt), with results validating the effectiveness of EvoStealer.

\section{Related Work}

We discuss two lines of related work: the text-to-image prompt stealing attacks and the evolutionary algorithms in LLMs.

\subsection{Text-to-Image Prompt Stealing Attack}
Prompt stealing attacks, or prompt extraction attacks, aim to infer the input from a model's output. A successful attack infringes on intellectual property and poses significant risks to prompt trading platforms in the era of LLMs. However, such attacks are more challenging in text-to-image generation due to the greater uncertainty in image generation compared to text. CLIP Interrogator employs CLIP~\citep{radford2021learning} to extract the subject and then selects phrases matching the target image from predefined sets~\citep{udo2023image}. ~\citet{shen2024prompt} fine-tunes 2 models to extract image subjects and modifiers separately, combining them for the attack. Building on this,~\citet{naseh2024iteratively} employ GPT-4V to iteratively optimize the prompt, resulting in higher quality. Unlike these works, EvoStealer targets the extraction of a generalizable prompt template, offering greater practical value compared to stealing individual prompts.

\subsection{Evolutionary Algorithms in LLMs}

Recent researches combining evolutionary algorithms with LLMs have demonstrated strong and stable performance across various tasks~\citep{yang2023large, liu2023large}. Some studies leverage the rich domain knowledge and powerful text analysis capabilities of LLMs to accelerate the search process in evolutionary algorithms, particularly in tasks involving complex reasoning~\citep{meyerson2024language,liu2024large,lange2024large,brahmachary2024large} and interpretability~\citep{chiquier2025evolving}. Conversely, some studies capitalize on the stability of evolutionary algorithms to utilize LLMs for generating higher-quality prompt words~\citep{xu2022gps,prasad2022grips,guo2023connecting,fernando2023promptbreeder}. In this paper, we use evolutionary algorithms to progressively generate style descriptors that closely resemble multiple example images, thereby achieving prompt template stealing.

\section{Data Consturction}

In this section, we introduce the threat model of prompt template stealing, providing a detailed description of the attacker's existing conditions, constraints, and objectives. We then detail our methodology for developing \data, a comprehensive benchmark designed to realistically simulate this attack scenario.  The specifics are presented below.

\subsection{Threat Model} \label{sec_threat_model}
The attack scenario is grounded in real-world applications. Attackers have access to two pieces of information from the prompt trading platform: 9 sample images and the generative model (e.g., DALL-E 3~\footnote{https://openai.com/index/dall-e-3/} or Midjourney~\footnote{https://www.midjourney.com/home}). While attackers can interact with the model via an API, they are not privy to its internal parameters. Their objectives are twofold: first, to generate images that closely resemble, or even replicate, the sample images by using the same subject with the stolen prompt template; and second, to alter the subject within the template and generate images that retain the same style as the sample images.

\subsection{Benchmark Construction} \label{sec_data_collection}
\begin{figure*}
    \centering
    \includegraphics[width=0.9\linewidth]{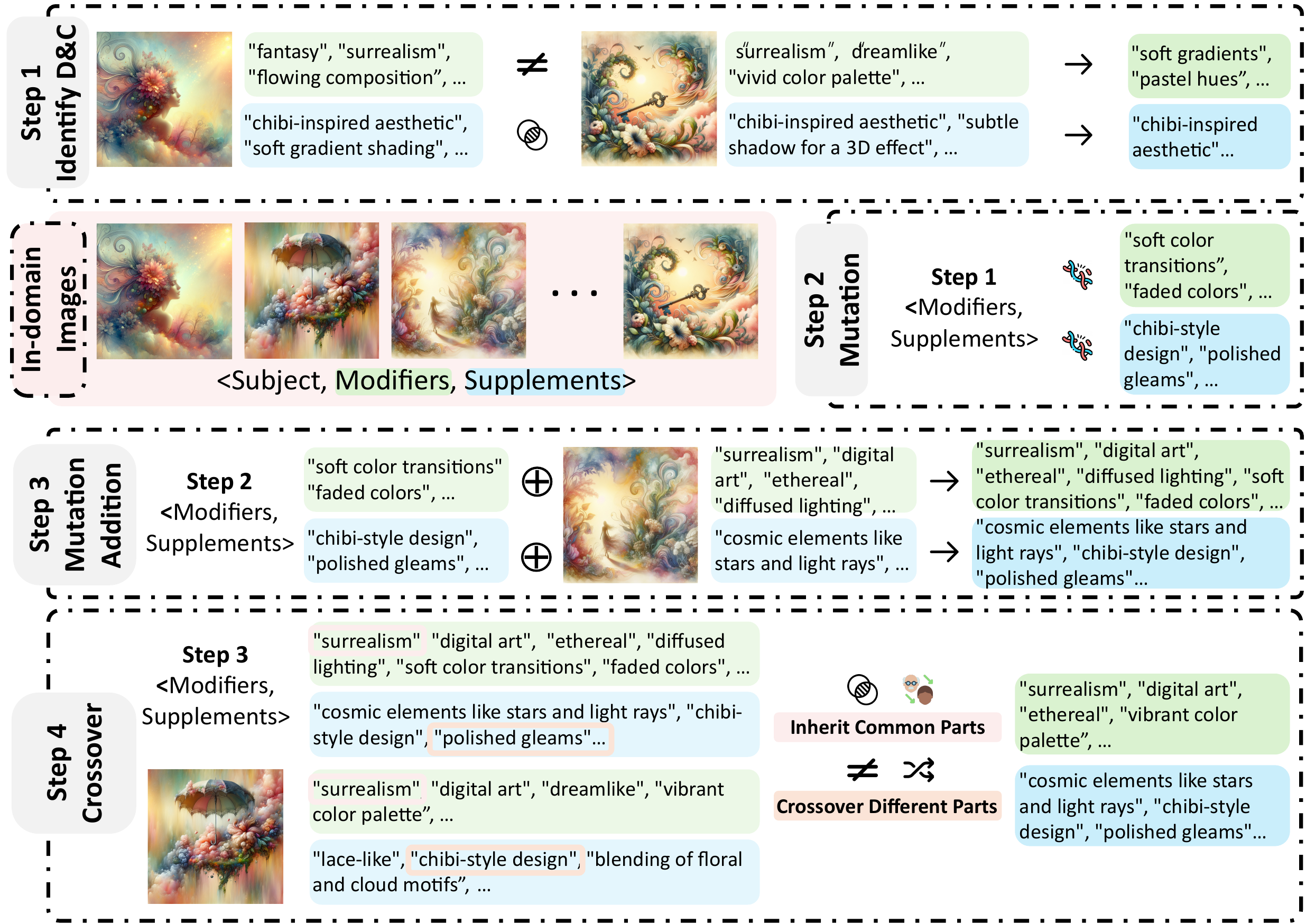}
    \caption{The key steps of EvoStealer in differential evolution, including the identification of differences and commonalities, mutation, mutation addition, and crossover operations.}
    \label{fig:framework}
\end{figure*}

Currently, no specialized benchmark exists for prompt template stealing research. To address this gap, we introduce \data, a novel benchmark comprising 50 freely available prompt templates sourced from PromptBase and LaPrompt. These templates are divided into two equal groups of 25 templates each, categorized as "Easy" and "Hard" based on complexity. Each group encompasses 9 distinct subject categories.
We utilize DALL·E 3 as our generation model, combining each prompt template with the 9 subjects to produce 450 unique images.
To ensure quality control, we implemented a comprehensive manual review process focusing on two key criteria: subject-prompt alignment and stylistic consistency across template-generated images. For each group, we designated the first 5 generated images as in-domain data to assess similarity between original and stolen prompt. The remaining 4 images serve as out-of-domain data, enabling evaluation of prompt template generalization capabilities across diverse subjects. For comprehensive details on the benchmark construction methodology, please refer to Appendix~\ref{app_data}.

\section{EvoStealer}
In this section, we introduce the three main steps of EvoStealer: Image Element Extraction, Differential Evolution, and Fitness Function. The details are presented below.

\subsection{Image Element Extraction}
High-quality prompts for text-to-image generation typically consist of a subject and several modifiers~\citep{liu2022design,oppenlaender2024taxonomy}. The subject defines the object or scene depicted in the image, such as \textit{"a woman with a flower crown"} or a more intricate description like \textit{"Woodland creatures gather around a shimmering pond, surrounded by trees and glowing flowers, creating a peaceful scene"}. The modifiers specify the style of the image, including aspects such as artistic style and resolution. While multimodal models can accurately identify simple subjects, they often misinterpret complex subjects, mistakenly treating parts of the subject as style modifiers. For instance, in the case of the complex subject "\textit{peaceful scene}", the model may misinterpret "\textit{peaceful}" as a style modifier, contaminating the intended description.

To address this issue, we define an image element extraction pattern: <Subject, Modifiers, Supplements>. The subject describes the object or scene, while the modifiers are categorized into four types: \textit{Artistic Style, Visual Composition and Structure, Aesthetic and Emotional Atmosphere}, and \textit{Medium and Material}. For further details, please refer to Appendix~\ref{app_pattern}. We have imposed the aforementioned constraints on the modifiers to ensure that the model describes the image solely from the relevant perspectives within the four predefined categories. This restriction contributes to the stability and controllability of modifier extraction. However, such a constraint may limit the model’s ability to fully capture the diversity of style features. To mitigate this limitation, we incorporate supplements as a compensatory measure. Supplements encompass descriptions outside the four categories and can include individual words, phrases, or even sentences, such as "\textit{radiating lines suggesting motion}" or "\textit{subtle transitions between colors}".

\begin{table*}[ht]
    \centering
    \small
    \setlength\tabcolsep{6.8pt} 
    \renewcommand{\arraystretch}{1}
    \scalebox{0.9}{ 
    \begin{tabular}{cccccc|c|c}
        \toprule
        \textbf{Method} & \textbf{DINO } & \( \textbf{CLIP}_{\textit{img}} \) & \( \textbf{CLIP}_{\textit{txt}} \) & \( \textbf{SigLIp}_{\textit{img}} \) & \( \textbf{SigLIp}_{\textit{txt}} \) & \color{purple}{\textbf{Average}} & \color{blue}{\textbf{Human Evaluation}}\\
        \midrule
        \multicolumn{8}{c}{\textit{Easy Benchmark}} \\ 
        \midrule\midrule
        BLIP 2~\cite{li2023blip}  & 62.07 & 79.38 & 48.35 & 82.32 & 52.69 & \color{purple}{64.96} & \color{blue}{3.42}\\
        CLIP Interrogator & 69.93 & 82.76 & 54.14 & 85.86 & 62.59 & \color{purple}{70.86} & \color{blue}{4.02}\\
        PromptStealer \cite{shen2024prompt}  & 63.73 & 77.90 & 49.21 & 82.73 & 61.93 & \color{purple}{67.10} & \color{blue}{3.78}\\\midrule
        EvoStealer (\intern)  & 74.68 & 84.46 & 68.94 & 87.88 & \textbf{74.93} & \color{purple}{78.18} & \color{blue}{4.32}\\
        EvoStealer (\gpt)  & 73.87 & 84.79 & 72.12 & 88.38 & 71.80 & \color{purple}{78.19} & \color{blue}{4.30}\\
        EvoStealer (\gpta)  & \textbf{75.83} & \textbf{85.30} & \textbf{74.41} & \textbf{89.14} & 72.75 & \color{purple}{\textbf{79.49}} & \color{blue}{4.52}\\
        \midrule
        \multicolumn{8}{c}{\textit{Hard Benchmark}} \\ 
        \midrule\midrule
        BLIP 2~\cite{li2023blip}  &61.16 & 76.67 & 46.04 & 80.51 & 50.74 & \color{purple}{63.02} & \color{blue}{3.24}\\
        CLIP Interrogator & 66.45 & 78.26 & 54.62 & 82.45 & 60.78 & \color{purple}{68.51} & \color{blue}{3.66}\\
        PromptStealer \cite{shen2024prompt}  & 60.01 & 75.58 & 47.10 & 79.20 & 59.71 & \color{purple}{64.32} & \color{blue}{3.48}\\\midrule
        EvoStealer (\intern)  & 70.16 & 80.63 & 63.02 & 84.66 & 68.14 & \color{purple}{73.32} & \color{blue}{4.17}\\
        EvoStealer (\gpt)  & \textbf{71.05} & 81.02 & 67.64 & 84.88 & 69.00 & \color{purple}{74.72} & \color{blue}{4.12}\\
        EvoStealer (\gpta)  & 69.24 & \textbf{81.34} & \textbf{70.61} & \textbf{85.28} & \textbf{69.27} & \color{purple}{\textbf{75.15}} & \color{blue}{4.24}\\

        \bottomrule
    \end{tabular}}
    \caption{The overall evaluation results for the in-domain data, with the bolded values indicating the best scores.}
    \label{tb:main_indomain}
\end{table*}
\subsection{Differential Evolution} \label{diffevo}
Firstly, we introduce the theory of differential evolution. The process of generating offspring through the differential evolution algorithm is represented using numerical vectors. Initially, each vector in the population is sequentially selected as the base vector, denoted as $\alpha$. Then, three individuals, $x_{1}, x_{2}, x_{3}$, are randomly chosen from the population to perform the mutation operation. Specifically, the difference between $x_{2}, x_{3}$ is calculated, and this difference undergoes mutation. The mutated difference is then added to $x_{1}$ to produce a new vector, denoted as $\beta$. The mutation operation is mathematically expressed as: $\beta = x_{1} + F(x_{2} - x_{3})$, where $F$ represents the mutation factor, which controls the magnitude of the mutation. Finally, a crossover operation is applied to the vectors $\alpha$ and $\beta$ to generate the offspring.

Figure~\ref{fig:framework} illustrates the differential evolution process implemented in EvoStealer. In Step 1, we differentiate between modifiers and supplements due to their distinct characteristics, particularly in terms of controllability and unpredictability. For modifiers, we focus on identifying the differences between the two sets, while for supplements, we concentrate on their common components. This approach is grounded in the understanding that the uncontrollability of supplements introduces unique features specific to individual images. Additionally, supplements typically contain more tokens than modifiers, which results in a greater influence on the visual representation of the image and, consequently, on the generalization ability of the template. In Step 2, we randomly select an image from the in-domain dataset to influence the mutation process. This strategy serves two purposes: first, it helps filter out modifiers that do not align with the image (e.g., in the case of a surrealistic style image, modifiers such as "cartoon style" are excluded); second, the image, serving as a mutation variable, introduces additional contextual information. As mentioned earlier, the initial version of EvoStealer directly derives image element extraction, which results in an over-reliance on the quality of this extraction. By incorporating the image in Step 2, we enable the population to gain valuable information that may otherwise be overlooked, thus mitigating the drawback of over-dependence on image element extraction. In Step 3, no modifications are made, and the two components are simply combined to generate the mutated description. In Step 4, in contrast to the direct crossover used in genetic algorithms, EvoStealer first identifies the common parts between the two individuals. When generating the new offspring, the common parts are fully inherited, while only the differing parts undergo the crossover operation. This design approach strikes a balance between the exploration and exploitation of the algorithm, facilitating effective exploration while ensuring that generalization constraints are preserved.

\subsection{Fitness Function} \label{section_fitness}
The fitness function is employed to assess the quality of offspring, with those exhibiting higher fitness scores being retained for progression to the next iteration. While the fitness function does not directly influence the offspring generation, it guides the search direction throughout the evolution process. Our fitness function incorporates both the semantic similarity of the text and the style similarity of the image. Specifically, for each offspring (i.e., a prompt template), we sequentially replace the subject within the template and calculate its semantic similarity with the ground truth. Additionally, we randomly select a subject and use the target model (DALL·E 3) to generate the corresponding image, subsequently calculating the similarity between this generated image and the corresponding image from the in-domain dataset. The mathematical formulation is as follows:

\[
\small
\begin{aligned}
F &= \frac{1}{n} \sum_{i=1}^{n} \left( \lambda \left( \frac{\mathbf{T}_{\text{off}}(i) \cdot \mathbf{I}_{\text{gt}}(i)}{\|\mathbf{T}_{\text{off}}(i)\| \|\mathbf{I}_{\text{gt}}(i)\|} \right) \right) \\
  &\quad + (1 - \lambda)  \left( \frac{\mathbf{I}_{\text{off}} \cdot \mathbf{I}_{\text{gt}}}{\|\mathbf{I}_{\text{off}}\| \|\mathbf{I}_{\text{gt}}\|} \right)
\end{aligned}
\]

Where \( \text{off} \) and \( \text{gt} \) denote the offspring and ground truth, respectively, and \( \mathbf{T} \) and \( \mathbf{I} \) represent the text and image embeddings. The parameter \(\lambda\) serves as a balance factor to weight the two similarity measures.

\section{Experiments}

We employ \data to evaluate the vulnerability of image generation models to prompt template stealing.
Following recent works~\citep{shen2024prompt, naseh2024iteratively, huang2024vbench}, we employ subject similarity, style similarity, and semantic similarity metrics to evaluate the performance of image generation models against prompt template stealing (Section \ref{metric}). These metrics demonstrate higher agreement with human annotations than previous approaches.
Additionally, we conduct human evaluation to measure the quality of prompt stealing.

\begin{table*}[ht]
    \centering
    \small
    \setlength\tabcolsep{7.2pt} 
    \renewcommand{\arraystretch}{1}
    \scalebox{0.9}{ 
    \begin{tabular}{cccccc|c|c}
        \toprule
        \textbf{Method} & \textbf{DINO } & \( \textbf{CLIP}_{\textit{img}} \) & \( \textbf{CLIP}_{\textit{txt}} \) & \( \textbf{SigLIp}_{\textit{img}} \) & \( \textbf{SigLIp}_{\textit{txt}} \) & \color{purple}{\textbf{Average}} & \color{blue}{\textbf{Human Evaluation}}\\
        \midrule
        \multicolumn{8}{c}{\textit{Easy Benchmark}} \\ 
        \midrule\midrule
        CLIP Interrogator  & 64.02 & 78.72 & 53.95 & 82.98 & 63.73 & \color{purple}{68.68} & \color{blue}{3.80}\\
        PromptStealer \cite{shen2024prompt}  & 60.53 & 75.53 & 51.37 & 81.19 & 61.16 & \color{purple}{65.96} & \color{blue}{3.64}\\\midrule
        EvoStealer (\intern)  & 72.93 & 83.13 & 68.63 & \textbf{87.24} & \textbf{74.54} & \color{purple}{77.29} & \color{blue}{4.36}\\        
        EvoStealer (\gpt)  & 74.53 & 83.60 & 71.87 & 85.28 & 73.30 & \color{purple}{77.71} & \color{blue}{4.47}\\
        
        EvoStealer (\gpta)  & \textbf{75.14} & \textbf{83.91} & \textbf{74.18} & 85.75 & 73.53 & \color{purple}{\textbf{79.10}} & \color{blue}{4.60}\\
        \midrule
        \multicolumn{8}{c}{\textit{Hard Benchmark}} \\ 
        \midrule\midrule
        CLIP Interrogator  & 62.23 & 69.90 & 51.66 & 75.19 & 58.51 & \color{purple}{63.50} & \color{blue}{3.74}\\
        PromptStealer \cite{shen2024prompt}  & 58.53 & 70.42 & 45.29 & 74.38 & 55.07 & \color{purple}{60.74} & \color{blue}{3.46}\\\midrule

        EvoStealer (\intern)  & 68.92 & 78.96 & 61.29 & 83.37 & 67.87 & \color{purple}{72.08} & \color{blue}{4.32}\\
        EvoStealer (\gpt)  & 67.76 & 79.55 & 66.91 & 84.13 & 68.84 & \color{purple}{73.44} & \color{blue}{4.35}\\
        EvoStealer (\gpta)  & \textbf{67.00} & \textbf{80.50} & \textbf{69.27} & \textbf{84.55} & \textbf{69.79} & \color{purple}{\textbf{74.22}} & \color{blue}{4.48}\\

        \bottomrule
    \end{tabular}}
    \caption{The overall evaluation results for the out-of-domain data, with the bolded values indicating the best scores.}
    \label{tb:main_outdomain}
\end{table*}

\subsection{Baselines}
Our baselines encompass models for both caption generation (BLIP-2) and prompt stealing attack~(CLIP Interrogator and PromptStealer).
\begin{itemize}
    \item \textbf{BLIP-2}: BLIP-2~\citep{li2023blip} is a multimodal model that aligns text with images using a lightweight Querying Transformer to connect a frozen image encoder with LLMs. In this study, we employ the BLIP-2-opt-2.7b model to generate image descriptions.
    \item \textbf{CLIP Interrogator}: CLIP Interrogator~\footnote{https://github.com/pharmapsychotic/clip-interrogator/tree/main} uses CLIP to generate image descriptions, incorporating prompts from preset categories such as artists, flavors, and mediums. It encodes both the image and text with the CLIP model, calculates their similarity, and generates the most matching description.
    \item \textbf{PromptStealer}: PromptStealer~\citep{shen2024prompt} consists of two modules: the Subject Generator, fine-tuned on BLIP to extract image subjects, and the Modifier Detector, a multi-class classifier that selects style modifiers based on similarity to predefined categories. The final prompt is generated by concatenating the subject and selected modifiers.
\end{itemize}

\subsection{Experimental Settings}

Due to the inherent difficulties in subject identification and replacement within BLIP-2-generated prompts, its evaluation is limited to in-domain data only. For both CLIP Interrogator and PromptStealer methods, we first extract subjects and modifiers from 5 in-domain samples and concatenate them to create prompts. We then randomly select a prompt and systematically replace its subject with subjects from the out-of-domain group. For PromptStealer, we maintain a threshold value of 0.6. In EvoStealer's implementation, we extract prompt templates from in-domain data and perform sequential subject substitutions, using 9 different subjects to generate the final prompts. Both the population size and generation count are set to 5, with the temperature parameter set to 0 to ensure consistent results. We employ SigLIP~\citep{zhai2023sigmoid} for fitness score calculations and set \(\lambda\) to 0.5. All image generation is performed using DALL·E 3 with a resolution of 1024×1024 and standard quality settings.

\subsection{Evaluation Metric} \label{metric}
We adopt the evaluation framework proposed by \citet{huang2024vbench} and employ the following metrics to assess the performance of EvoStealer and baseline methods:
\begin{itemize} 
    \item \textbf{Subject Similarity}: To evaluate the similarity between subjects in paired images, we utilize the self-supervised model DINO~\citep{oquab2023dinov2}, as subject comparison is a crucial aspect of image similarity assessment.
    \item \textbf{Style Similarity}: To measure style consistency, we employ CLIP and SigLIP to extract style features from images generated using stolen prompts and compare them with the original images.
    \item \textbf{Semantic Similarity}: To assess prompt similarity, we compute the cosine similarity between embeddings of the stolen and target prompts, generated using CLIP and SigLIP.
    \item \textbf{Human Evaluation}: We recruit 3 external evaluators to rate the similarity between generated and target images on a scale of 1-5, where higher scores indicate greater similarity. For each group, we randomly sample 2 images from both in-domain and out-of-domain categories and calculate average scores. For out-of-domain samples, the evaluation focuses exclusively on style similarity. For further details, please
    refer to Appendix~\ref{app_he}. 
\end{itemize}

\subsection{Main Results}

Tables~\ref{tb:main_indomain} and~\ref{tb:main_outdomain} present comparative performance evaluations between EvoStealer and baseline methods using both in-domain and out-of-domain data. The results demonstrate that EvoStealer consistently outperforms baseline approaches across all evaluation metrics.

\paragraph{Performance on in-domain data.} EvoStealer outperforms other methods in both the Easy and Hard categories. For example, EvoStealer (GPT-4o) leads the second-best method, CLIP Interrogator, by 8.63\% and 6.64\% on the two datasets, demonstrating its ability to generate more accurate prompt templates and better stealing performance. Notably, EvoStealer excels in textual semantic comparison, as its prompts are significantly more effective. CLIP Interrogator and PromptStealer rely on simple concatenation of [subject] and modifiers, limiting variability. Additionally, CLIP’s length restriction hampers modifier extraction. In contrast, EvoStealer generates diverse templates iteratively, avoiding these limitations. This aligns with the findings of~\citet{naseh2024iteratively}.

\paragraph{Performance on out-of-domain data.} As shown in Table~\ref{tb:main_outdomain}, EvoStealer outperforms other methods, especially on out-of-domain data, where it demonstrates a larger advantage compared to in-domain data. For instance, EvoStealer (GPT-4o) leads by more than 10\% across all data types, indicating better generalization of stolen templates to different subjects. Furthermore, as seen in Table~\ref{tb:main_indomain}, EvoStealer's performance on out-of-domain data shows minimal degradation, while CLIP Interrogator and PromptStealer experience average degradations of 3.60\% and 2.36\%, respectively. This is due to EvoStealer’s effective template stealing by extracting common features across multiple images.

\paragraph{Comparison of performance across different models.} As shown in Tables~\ref{tb:main_indomain} and~\ref{tb:main_outdomain}, GPT-4o outperforms the other models, followed by GPT-4o-mini, with InternVL2-26B performing slightly worse. However, the performance differences among these models are minimal. This is primarily due to EvoStealer's reliance on the models' text and image analysis capabilities, indicating that EvoStealer is highly compatible and not dependent on a specific multimodal model.
\section{Analysis}

In this section, we analyze the effects of EvoStealer's components, the iteration number, and the experimental costs.

\subsection{Ablation Study} \label{ablation}

\begin{table}[!t]
    \centering
    \small
    \setlength\tabcolsep{8.8pt} 
    \renewcommand{\arraystretch}{0.9}  

    \begin{tabular}{ccc|c}
        \toprule
        \textbf{Method} & \textbf{InDom.} & \textbf{OutDom.} & \textbf{Average} \\
        \midrule
        Ours  & 77.32 & 76.67 & 77.00 \\ 
        \hspace{2pt}\textit{w/o.} supp.   & 73.56 & 74.85 & 74.21 \\
        \hspace{2pt}\textit{w/o.} img.    & 75.89 & 75.57 & 75.73 \\ 
        \bottomrule
    \end{tabular}
    \caption{Results of the ablation study: Impact of omitting supplements in the extraction pattern (w/o supp.) and excluding image similarity in the fitness function (w/o img.), with the model employed being GPT-4.}
    \label{tb:abalation}
\end{table}

We remove the supplements from the extracted templates and the image similarity evaluation from the fitness function to examine their impact on EvoStealer. The results are shown in Table~\ref{tb:abalation}. As observed, removing either module results in decreased performance, with a more significant drop when supplements are removed—specifically, an average similarity reduction of 2.79\%. This is because supplements provide additional details, such as image features and style information. As noted in Section~\ref{diffevo}, supplements are longer than individual modifiers, so their removal has a more pronounced effect on visual performance. A comparison of the performance before and after removing supplements is provided in Appendix~\ref{app_ablation}. Removing the image similarity evaluation from the fitness function causes a performance decrease of 1.27\%, suggesting that including the comparison between the generated and target images in the fitness function helps guide the evolutionary process and accelerate convergence.
\begin{figure}[h!]
    \centering
    \includegraphics[width=\linewidth]{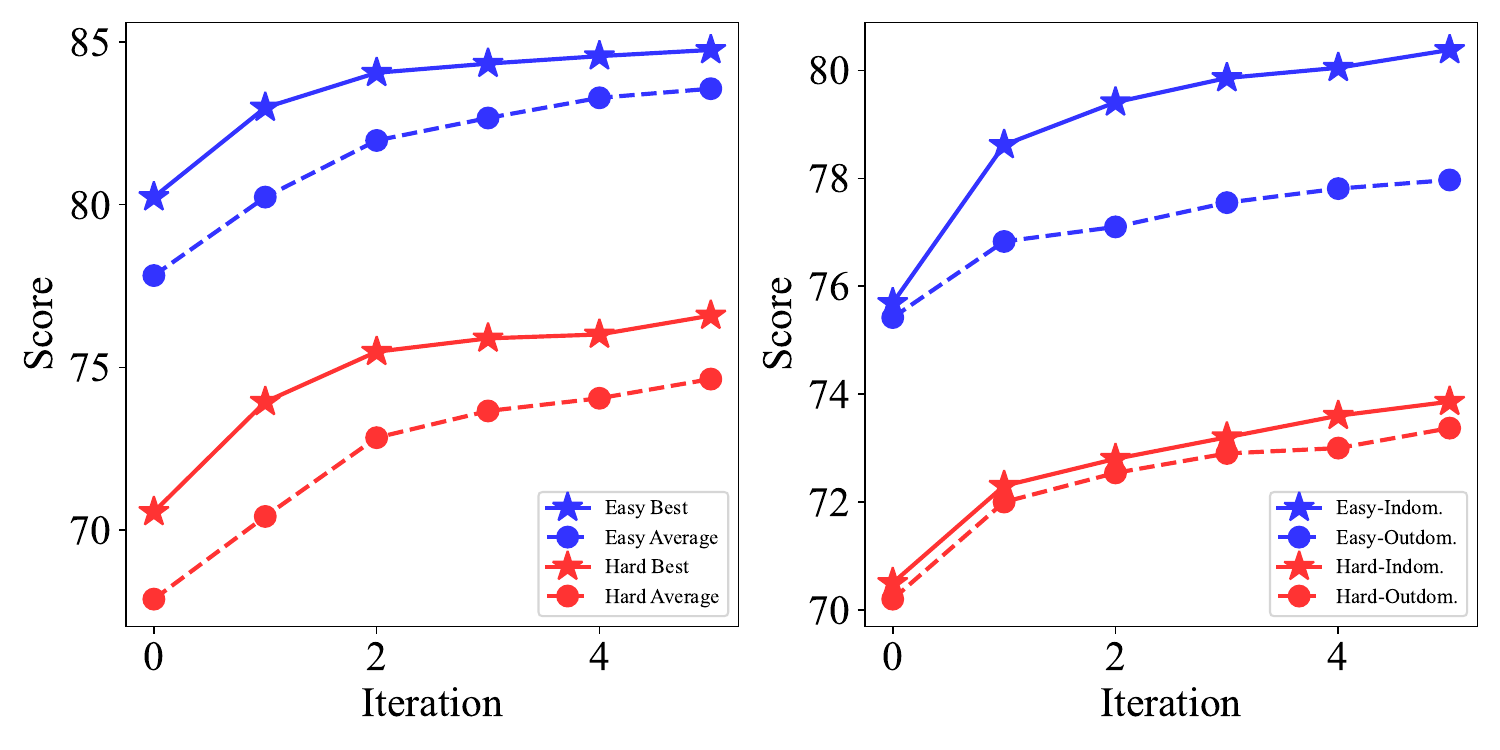}
    \caption{The convergence curve of EvoStealer, with the left half showing changes in fitness score and the right half depicting performance changes of the optimal prompt template for in-domain and out-of-domain data.}
    \label{fig:iters}
\end{figure}

\begin{figure*}[h!]
    \centering
    \includegraphics[width=0.9\linewidth]{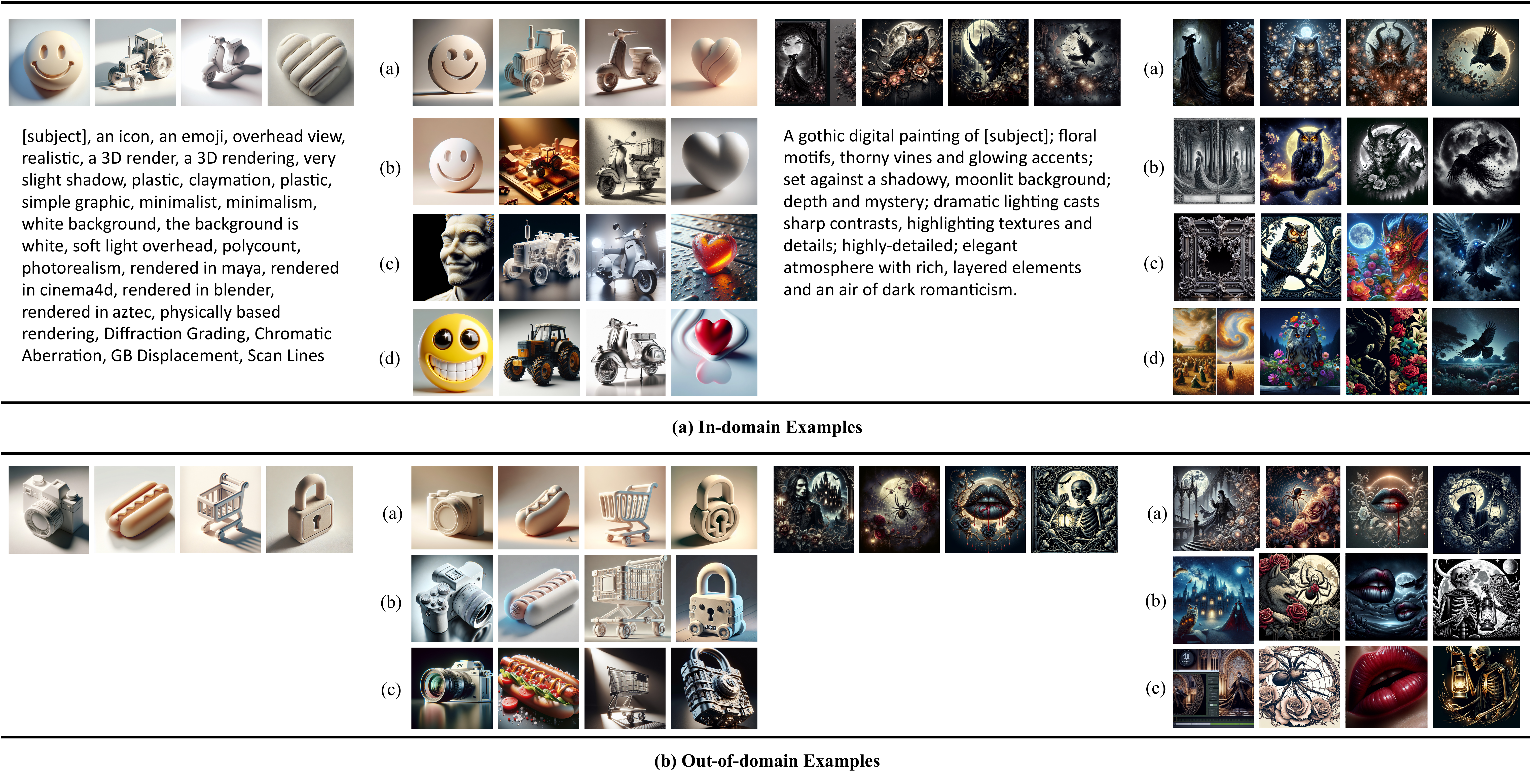}
    \caption{The attack results of EvoStealer compared to three baseline methods on both easy and hard examples. (a)-(d) represent EvoStealer, CLIP-Interrogator, PromptStealer, and BLIP2, respectively.}
    \label{fig:case}
\end{figure*}
\subsection{Effect of Number of Iterations}

We select 10 groups of easy and 10 groups of hard cases to examine EvoStealer's convergence~(we use GPT-4o as the analysis model), with results shown in Figure~\ref{fig:iters}. The left section of the figure displays changes in the fitness score as evolution progresses, while the right section shows changes in the scores of the optimal templates for both in-domain and out-of-domain data. We observe that as evolution progresses, both the optimal and average fitness scores gradually increase, indicating that EvoStealer generates offspring with higher adaptability. The performance of the prompt templates steadily improves for both in-domain and out-of-domain data. Two examples are provided in Appendix~\ref{app_progress}

\subsection{Cost Analysis}

To assess the practicality of EvoStealer, we analyzed the cost of stealing a prompt template. The primary overhead of EvoStealer consists of three components: population initialization, differential evolution (including the fitness function), and image synthesis. A detailed cost estimation process is provided in Appendix~\ref{app_cost}. The results indicate that EvoStealer requires 144 API calls, generates 34 images (including 9 final synthesized images), and consumes approximately 119.1k tokens, amounting to a total cost of \$1.70. While this is lower than the platform’s pricing range of \$3–9, the cost advantage is not substantial. However, as demonstrated in the ablation study in Section~\ref{ablation}, costs can be further reduced by using open-source models or omitting image similarity calculations in the fitness function, enabling near-zero-cost stealing. Although this cost-reduced version performs slightly worse than the full EvoStealer model, it still significantly outperforms alternative approaches.

\section{Case Study}
To clearly demonstrate EvoStealer's advantages over baseline methods, we select an easy and a hard example for case study, with the results shown in Figure~\ref{fig:case}. The results show that, on in-domain data, EvoStealer generates images that closely match the style of the original images, with all four synthesized images maintaining stylistic consistency. In contrast, the four images generated by the other baseline methods exhibit significant style variation. On out-of-domain data, EvoStealer maintains the same style as in-domain images, successfully achieving subject generalization. In contrast, the other baseline methods fail to generalize. Additionally, we analyze three distinct failure cases (see Appendix~\ref{app_failcases} for details).
\section{Conclusion}
This paper investigates prompt template stealing—whether attackers can extract generalizable templates that maintain stylistic consistency using minimal sample images. To explore this scenario, we provide \data, a two-tier benchmark consisting of 50 templates and 450 images, organized into Easy and Hard difficulty levels. We also introduce EvoStealer, a template stealing method that combines differential evolution algorithms with MLLMs, enabling template stealing without the need for fine-tuning. Extensive experiments and analysis validate its effectiveness and practicality.

\section{Limitations}
The current implementation of EvoStealer and benchmark presents several methodological limitations:
\begin{enumerate}
    \item EvoStealer's MLLM-based design offers simplified implementation without fine-tuning requirements and maintains robust performance across open datasets. However, this approach inherently limits the system's maximum performance to the capabilities of the underlying MLLMs.
    \item Resource constraints restricted our benchmark to DALL·E-3 generated images, excluding other prominent models like Midjourney and Stable Diffusion. Nevertheless, the current benchmark adequately evaluates stealing method performance, with planned expansion to additional models in future work.
    \item The benchmark's single-subject design facilitates comparative analysis but does not address multi-subject templates in real-world applications—a limitation to be addressed in subsequent research.
\end{enumerate}

\section{Ethical Considerations}
EvoStealer's ability to extract prompt templates from minimal image examples enables attackers to generate multiple stylistically similar images through minor template modifications, posing significant risks to creators' intellectual property. This research highlights this security vulnerability, as understanding such threat models is essential for developing effective countermeasures.

While watermarking offers some protection, its implementation on trading platforms presents practical challenges. Watermarks can obscure image details, potentially deterring buyers or leading to customer dissatisfaction when purchased prompts fail to meet expectations. Our findings suggest that limiting the number of displayed images to 2-4 examples provides a simple yet effective defensive strategy.

Future research should prioritize developing robust protection mechanisms to safeguard both creators' rights and the integrity of the AI-generated content marketplace.

\bibliography{custom}

\begin{thebibliography}{29}
\providecommand{\natexlab}[1]{#1}

\bibitem[{Brahmachary et~al.(2024)Brahmachary, Joshi, Panda, Koneripalli, Sagotra, Patel, Sharma, Jagtap, and Kalyanaraman}]{brahmachary2024large}
Shuvayan Brahmachary, Subodh~M Joshi, Aniruddha Panda, Kaushik Koneripalli, Arun~Kumar Sagotra, Harshil Patel, Ankush Sharma, Ameya~D Jagtap, and Kaushic Kalyanaraman. 2024.
\newblock Large language model-based evolutionary optimizer: Reasoning with elitism.
\newblock \emph{arXiv preprint arXiv:2403.02054}.

\bibitem[{Cao et~al.(2024)Cao, Zhou, Song, and Yang}]{cao2024controllable}
Pu~Cao, Feng Zhou, Qing Song, and Lu~Yang. 2024.
\newblock Controllable generation with text-to-image diffusion models: A survey.
\newblock \emph{arXiv preprint arXiv:2403.04279}.

\bibitem[{Chiquier et~al.(2025)Chiquier, Mall, and Vondrick}]{chiquier2025evolving}
Mia Chiquier, Utkarsh Mall, and Carl Vondrick. 2025.
\newblock Evolving interpretable visual classifiers with large language models.
\newblock In \emph{European Conference on Computer Vision}, pages 183--201. Springer.

\bibitem[{Fernando et~al.(2023)Fernando, Banarse, Michalewski, Osindero, and Rockt{\"a}schel}]{fernando2023promptbreeder}
Chrisantha Fernando, Dylan Banarse, Henryk Michalewski, Simon Osindero, and Tim Rockt{\"a}schel. 2023.
\newblock Promptbreeder: Self-referential self-improvement via prompt evolution.
\newblock \emph{arXiv preprint arXiv:2309.16797}.

\bibitem[{Guo et~al.(2023)Guo, Wang, Guo, Li, Song, Tan, Liu, Bian, and Yang}]{guo2023connecting}
Qingyan Guo, Rui Wang, Junliang Guo, Bei Li, Kaitao Song, Xu~Tan, Guoqing Liu, Jiang Bian, and Yujiu Yang. 2023.
\newblock Connecting large language models with evolutionary algorithms yields powerful prompt optimizers.
\newblock \emph{arXiv preprint arXiv:2309.08532}.

\bibitem[{Ho et~al.(2020)Ho, Jain, and Abbeel}]{ho2020denoising}
Jonathan Ho, Ajay Jain, and Pieter Abbeel. 2020.
\newblock Denoising diffusion probabilistic models.
\newblock \emph{Advances in neural information processing systems}, 33:6840--6851.

\bibitem[{Huang et~al.(2024)Huang, He, Yu, Zhang, Si, Jiang, Zhang, Wu, Jin, Chanpaisit et~al.}]{huang2024vbench}
Ziqi Huang, Yinan He, Jiashuo Yu, Fan Zhang, Chenyang Si, Yuming Jiang, Yuanhan Zhang, Tianxing Wu, Qingyang Jin, Nattapol Chanpaisit, et~al. 2024.
\newblock Vbench: Comprehensive benchmark suite for video generative models.
\newblock In \emph{Proceedings of the IEEE/CVF Conference on Computer Vision and Pattern Recognition}, pages 21807--21818.

\bibitem[{Lange et~al.(2024)Lange, Tian, and Tang}]{lange2024large}
Robert Lange, Yingtao Tian, and Yujin Tang. 2024.
\newblock Large language models as evolution strategies.
\newblock In \emph{Proceedings of the Genetic and Evolutionary Computation Conference Companion}, pages 579--582.

\bibitem[{Li et~al.(2023)Li, Li, Savarese, and Hoi}]{li2023blip}
Junnan Li, Dongxu Li, Silvio Savarese, and Steven Hoi. 2023.
\newblock Blip-2: Bootstrapping language-image pre-training with frozen image encoders and large language models.
\newblock In \emph{International conference on machine learning}, pages 19730--19742. PMLR.

\bibitem[{Liu et~al.(2023)Liu, Lin, Wang, Yao, Tong, Yuan, and Zhang}]{liu2023large}
Fei Liu, Xi~Lin, Zhenkun Wang, Shunyu Yao, Xialiang Tong, Mingxuan Yuan, and Qingfu Zhang. 2023.
\newblock Large language model for multi-objective evolutionary optimization.
\newblock \emph{arXiv preprint arXiv:2310.12541}.

\bibitem[{Liu et~al.(2024{\natexlab{a}})Liu, Li, Wu, and Lee}]{liu2024visual}
Haotian Liu, Chunyuan Li, Qingyang Wu, and Yong~Jae Lee. 2024{\natexlab{a}}.
\newblock Visual instruction tuning.
\newblock \emph{Advances in neural information processing systems}, 36.

\bibitem[{Liu et~al.(2024{\natexlab{b}})Liu, Chen, Qu, Tang, and Ong}]{liu2024large}
Shengcai Liu, Caishun Chen, Xinghua Qu, Ke~Tang, and Yew-Soon Ong. 2024{\natexlab{b}}.
\newblock Large language models as evolutionary optimizers.
\newblock In \emph{2024 IEEE Congress on Evolutionary Computation (CEC)}, pages 1--8. IEEE.

\bibitem[{Liu and Chilton(2022)}]{liu2022design}
Vivian Liu and Lydia~B Chilton. 2022.
\newblock Design guidelines for prompt engineering text-to-image generative models.
\newblock In \emph{Proceedings of the 2022 CHI conference on human factors in computing systems}, pages 1--23.

\bibitem[{Meyerson et~al.(2024)Meyerson, Nelson, Bradley, Gaier, Moradi, Hoover, and Lehman}]{meyerson2024language}
Elliot Meyerson, Mark~J Nelson, Herbie Bradley, Adam Gaier, Arash Moradi, Amy~K Hoover, and Joel Lehman. 2024.
\newblock Language model crossover: Variation through few-shot prompting.
\newblock \emph{ACM Transactions on Evolutionary Learning}, 4(4):1--40.

\bibitem[{Naseh et~al.(2024)Naseh, Thai, Iyyer, and Houmansadr}]{naseh2024iteratively}
Ali Naseh, Katherine Thai, Mohit Iyyer, and Amir Houmansadr. 2024.
\newblock Iteratively prompting multimodal llms to reproduce natural and ai-generated images.
\newblock \emph{arXiv preprint arXiv:2404.13784}.

\bibitem[{Oppenlaender(2024)}]{oppenlaender2024taxonomy}
Jonas Oppenlaender. 2024.
\newblock A taxonomy of prompt modifiers for text-to-image generation.
\newblock \emph{Behaviour \& Information Technology}, 43(15):3763--3776.

\bibitem[{Oquab et~al.(2023)Oquab, Darcet, Moutakanni, Vo, Szafraniec, Khalidov, Fernandez, Haziza, Massa, El-Nouby et~al.}]{oquab2023dinov2}
Maxime Oquab, Timoth{\'e}e Darcet, Th{\'e}o Moutakanni, Huy Vo, Marc Szafraniec, Vasil Khalidov, Pierre Fernandez, Daniel Haziza, Francisco Massa, Alaaeldin El-Nouby, et~al. 2023.
\newblock Dinov2: Learning robust visual features without supervision.
\newblock \emph{arXiv preprint arXiv:2304.07193}.

\bibitem[{Prasad et~al.(2022)Prasad, Hase, Zhou, and Bansal}]{prasad2022grips}
Archiki Prasad, Peter Hase, Xiang Zhou, and Mohit Bansal. 2022.
\newblock Grips: Gradient-free, edit-based instruction search for prompting large language models.
\newblock \emph{arXiv preprint arXiv:2203.07281}.

\bibitem[{Radford et~al.(2021)Radford, Kim, Hallacy, Ramesh, Goh, Agarwal, Sastry, Askell, Mishkin, Clark et~al.}]{radford2021learning}
Alec Radford, Jong~Wook Kim, Chris Hallacy, Aditya Ramesh, Gabriel Goh, Sandhini Agarwal, Girish Sastry, Amanda Askell, Pamela Mishkin, Jack Clark, et~al. 2021.
\newblock Learning transferable visual models from natural language supervision.
\newblock In \emph{International conference on machine learning}, pages 8748--8763. PMLR.

\bibitem[{Sha and Zhang(2024)}]{sha2024prompt}
Zeyang Sha and Yang Zhang. 2024.
\newblock Prompt stealing attacks against large language models.
\newblock \emph{arXiv preprint arXiv:2402.12959}.

\bibitem[{Shen et~al.(2024)Shen, Qu, Backes, and Zhang}]{shen2024prompt}
Xinyue Shen, Yiting Qu, Michael Backes, and Yang Zhang. 2024.
\newblock Prompt stealing attacks against $\{$Text-to-Image$\}$ generation models.
\newblock In \emph{33rd USENIX Security Symposium (USENIX Security 24)}, pages 5823--5840.

\bibitem[{Sohl-Dickstein et~al.(2015)Sohl-Dickstein, Weiss, Maheswaranathan, and Ganguli}]{sohl2015deep}
Jascha Sohl-Dickstein, Eric Weiss, Niru Maheswaranathan, and Surya Ganguli. 2015.
\newblock Deep unsupervised learning using nonequilibrium thermodynamics.
\newblock In \emph{International conference on machine learning}, pages 2256--2265. PMLR.

\bibitem[{Udo and Koshinaka(2023)}]{udo2023image}
Honori Udo and Takafumi Koshinaka. 2023.
\newblock Image captioners sometimes tell more than images they see.
\newblock \emph{arXiv preprint arXiv:2305.02932}.

\bibitem[{Wang et~al.(2024)Wang, Bai, Tan, Wang, Fan, Bai, Chen, Liu, Wang, Ge et~al.}]{wang2024qwen2}
Peng Wang, Shuai Bai, Sinan Tan, Shijie Wang, Zhihao Fan, Jinze Bai, Keqin Chen, Xuejing Liu, Jialin Wang, Wenbin Ge, et~al. 2024.
\newblock Qwen2-vl: Enhancing vision-language model's perception of the world at any resolution.
\newblock \emph{arXiv preprint arXiv:2409.12191}.

\bibitem[{Wu et~al.(2024)Wu, Gao, Zhu, Zhou, Sun, Yang, Lou, Ding, and Yang}]{wu2024strago}
Yurong Wu, Yan Gao, Bin~Benjamin Zhu, Zineng Zhou, Xiaodi Sun, Sheng Yang, Jian-Guang Lou, Zhiming Ding, and Linjun Yang. 2024.
\newblock Strago: Harnessing strategic guidance for prompt optimization.
\newblock \emph{arXiv preprint arXiv:2410.08601}.

\bibitem[{Xu et~al.(2022)Xu, Chen, Du, Shao, Wang, Li, and Yang}]{xu2022gps}
Hanwei Xu, Yujun Chen, Yulun Du, Nan Shao, Yanggang Wang, Haiyu Li, and Zhilin Yang. 2022.
\newblock Gps: Genetic prompt search for efficient few-shot learning.
\newblock \emph{arXiv preprint arXiv:2210.17041}.

\bibitem[{Yang et~al.(2023)Yang, Wang, Lu, Liu, Le, Zhou, and Chen}]{yang2023large}
Chengrun Yang, Xuezhi Wang, Yifeng Lu, Hanxiao Liu, Quoc~V Le, Denny Zhou, and Xinyun Chen. 2023.
\newblock Large language models as optimizers.
\newblock \emph{arXiv preprint arXiv:2309.03409}.

\bibitem[{Yang et~al.(2024)Yang, Wu, Gao, Zhou, Zhu, Sun, Lou, Ding, Hu, Fang et~al.}]{yang2024ampo}
Sheng Yang, Yurong Wu, Yan Gao, Zineng Zhou, Bin~Benjamin Zhu, Xiaodi Sun, Jian-Guang Lou, Zhiming Ding, Anbang Hu, Yuan Fang, et~al. 2024.
\newblock Ampo: Automatic multi-branched prompt optimization.
\newblock \emph{arXiv preprint arXiv:2410.08696}.

\bibitem[{Zhai et~al.(2023)Zhai, Mustafa, Kolesnikov, and Beyer}]{zhai2023sigmoid}
Xiaohua Zhai, Basil Mustafa, Alexander Kolesnikov, and Lucas Beyer. 2023.
\newblock Sigmoid loss for language image pre-training.
\newblock In \emph{Proceedings of the IEEE/CVF International Conference on Computer Vision}, pages 11975--11986.

\end{thebibliography}

\appendix
\section{Data Collection} \label{app_data}
\begin{figure}[h!]
    \centering
    \includegraphics[width=\linewidth]{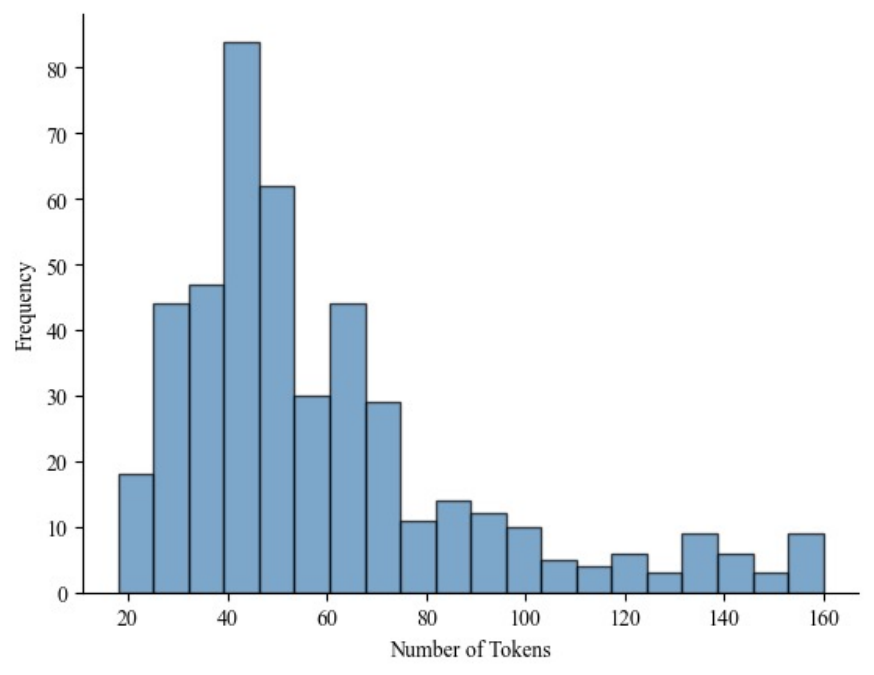}
    \caption{Token frequency distribution of the dataset}
    \label{fig:data}
\end{figure}

Our data collection and preprocessing pipeline consists of several systematic steps. Initially, we collect 100 free templates (excluding specific images) from PromptBase and LaPrompt, subsequently generating 900 corresponding images using DALL-E 3. Through manual curation, we eliminate templates that prove challenging to reproduce, such as those containing specific artistic style descriptors (e.g., Arkhip Kuindzhi). We then perform deduplication to remove images with highly similar styles and subjects, thereby preventing evaluation bias from data redundancy.
The subsequent quality control process encompasses two primary aspects: subject alignment verification and style consistency assessment. When anomalous data is identified, we regenerate images using DALL-E 3 until they meet our quality criteria. Finally, we categorize our dataset into Easy and Hard classifications. The Hard category is characterized by: uncommon modifiers, abstract subject descriptions and rich image details. The token distribution of the complete dataset is illustrated in Figure~\ref{fig:data}, while Table~\ref{tb:data} presents a detailed breakdown of token statistics for both Easy and Hard categories.

\section{Extraction Pattern Detail} \label{app_pattern}
Describing the style of an image requires including different perspectives. The style description of EvoStealer includes four categories: \textit{Artistic Style, Visual Composition and Structure, Aesthetic and Emotional Atmosphere}, and \textit{Medium and Material}.
\begin{itemize}
    \item Artistic Style: Include Genre, Era or Historical Style, Cultural and Technological Style.
    \item Visual Composition and Structure: Include Composition and Layout, Form and Structure, Scale, Movement, Perspective, Pattern and Ornamentation and Detail Level.
    \item Aesthetic and Emotional Atmosphere: Include Tone and Atmosphere, Emotional Atmosphere, Lighting and Shadow Effects,
    \item Medium and Material: Include Medium, Material, Technique, Texture, Surface, Color Palette, Brushwork, Line Quality, Strokes, Layering, Transparency, Opacity and Resolution.
\end{itemize}

\begin{table}[!t]
    \centering
    \small
    \renewcommand{\arraystretch}{1}  
    \setlength{\tabcolsep}{5pt}
    \begin{tabular}{ccccc}
        \toprule
        & \multicolumn{2}{c}{\textbf{Easy}} & \multicolumn{2}{c}{\textbf{Hard}} \\
        \cmidrule{2-5}
        \multicolumn{1}{c}{\multirow{-2}{*}{\textbf{Number}}} & Subject & Modifier & Subject & Modifier \\
        \midrule
        Min. & 1 & 23 & 1 & 16\\
        Max. & 27 & 154 & 24 & 107\\
        Avg. &  8.03 & 65.00 & 3.60 & 43.64\\
        \bottomrule
    \end{tabular}
    \caption{Token Statistics for Easy and Hard Benchmarks.}
    \label{tb:data}
\end{table}

\section{Human Evaluation} \label{app_he}
We implement a rigorous human evaluation protocol using a blinded manual scoring approach. Each evaluator is presented with the original benchmark images alongside extracted results from all methods, comprising two in-domain and two out-of-domain images per set. To maintain objectivity, evaluators are blinded to the generation methods and conduct their assessments independently, without inter-evaluator communication. The evaluation criteria are differentiated by image category:
\begin{itemize}
    \item For in-domain data: Evaluators assess both subject matter and stylistic similarity to measure template reproduction fidelity
    \item For out-of-domain data: Evaluation focuses exclusively on stylistic similarity to assess template generalization capability
\end{itemize}
Images are rated using a 5-point Likert scale, with higher scores indicating greater similarity. Final results are reported as mean scores across all evaluators. The detailed scoring criteria are presented below.
\begin{enumerate}
    \item \textbf{Completely Different:} The generated image exhibits no discernible similarities to the original, presenting entirely distinct content and stylistic elements.
    \item \textbf{Barely Similar:} While minimal thematic or elemental commonalities may exist between the original and generated images, they demonstrate significant divergence in both content and stylistic execution.
    \item \textbf{Somewhat Similar:} The generated image maintains recognizable correspondence to the original's content or subject matter, although notable stylistic variations are present.
    \item \textbf{Closely Similar:} The generated image demonstrates substantial fidelity to the original's content and subject matter, with only minor compositional variations.
    \item \textbf{Very Similar:} The generated image achieves near-identical reproduction, maintaining high fidelity to the original's content, style, and intricate details.
\end{enumerate}

\section{Ablation Comparison} \label{app_ablation}
\newcolumntype{C}[1]{>{\centering\arraybackslash}m{#1}}

\begin{figure*}[t]
    \vspace{-10pt}
    \centering
    \begin{adjustbox}{width=\textwidth,center}
    \begin{tabular}{C{4.5cm}C{4.5cm}C{4.5cm}C{2cm}}
        \hline
        \textbf{Original Images} & \textbf{EvoStealer} & \textbf{EvoStealer (\textit{w/o.} supp.)} & \textbf{Note} \\
        \hline
        \\
        \raisebox{-0.05\height}{\includegraphics[width=4.4cm]{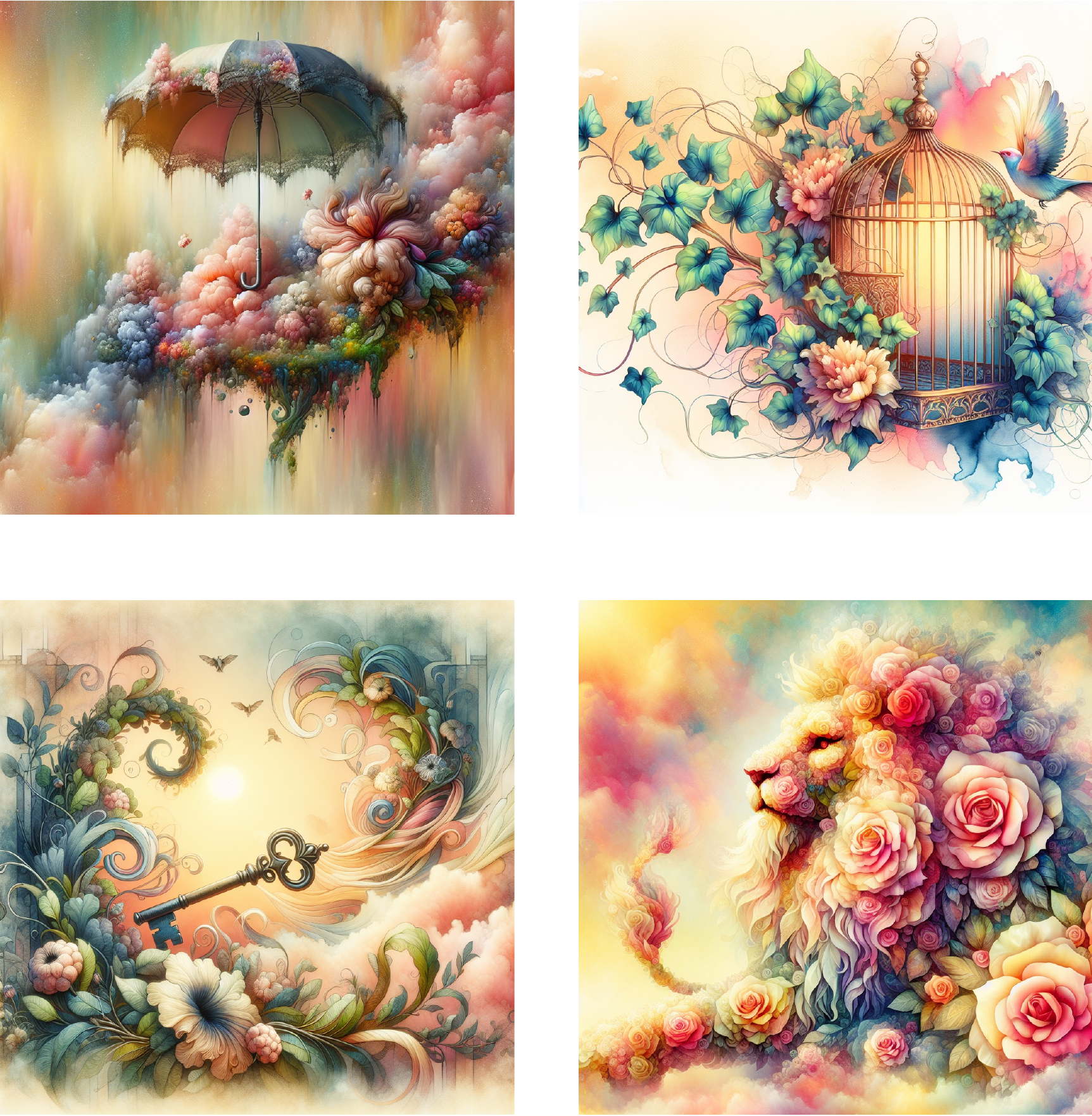}} & 
        \raisebox{-0.05\height}{\includegraphics[width=4.4cm]{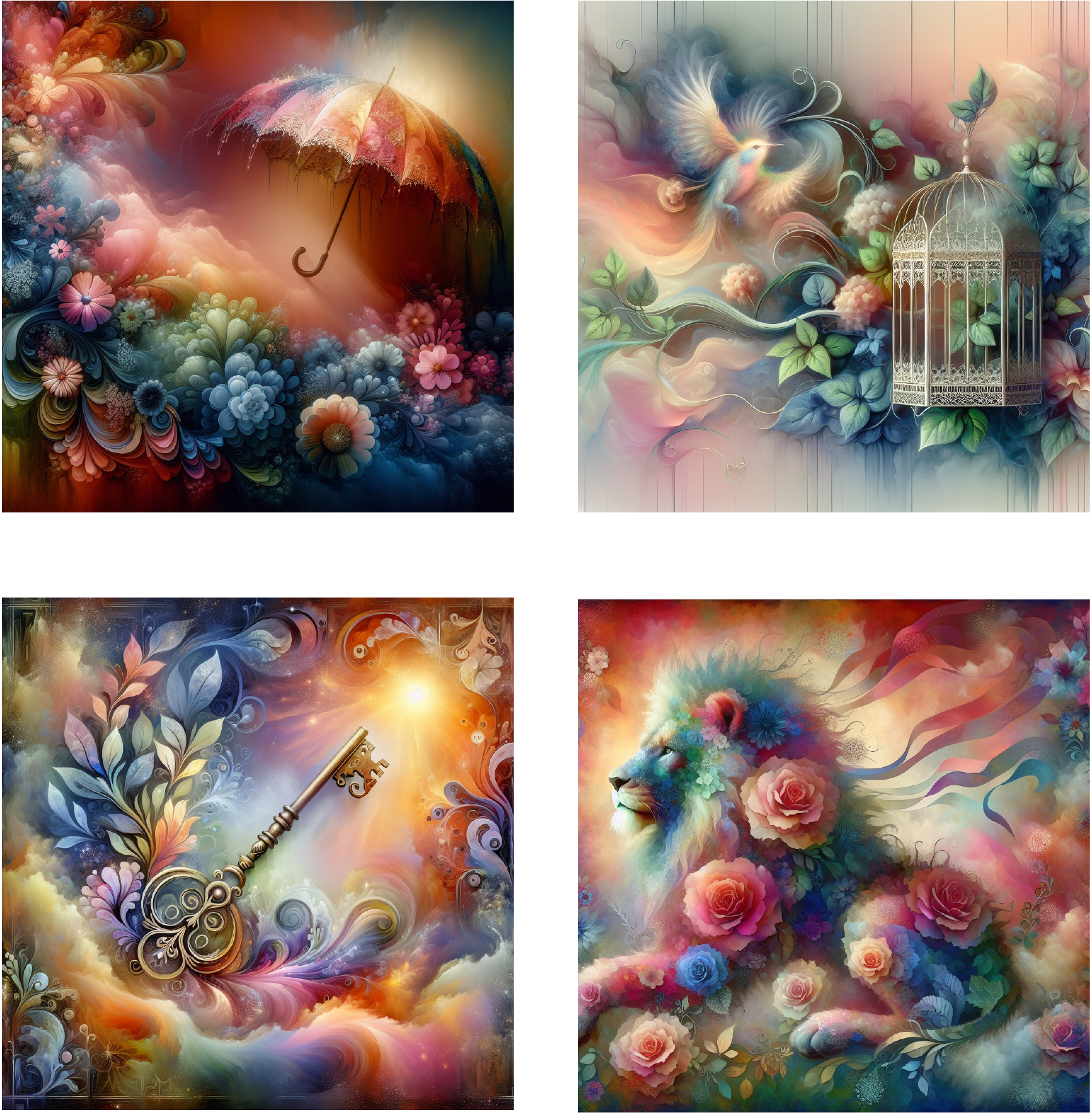}} &
        \raisebox{-0.05\height}{\includegraphics[width=4.4cm]{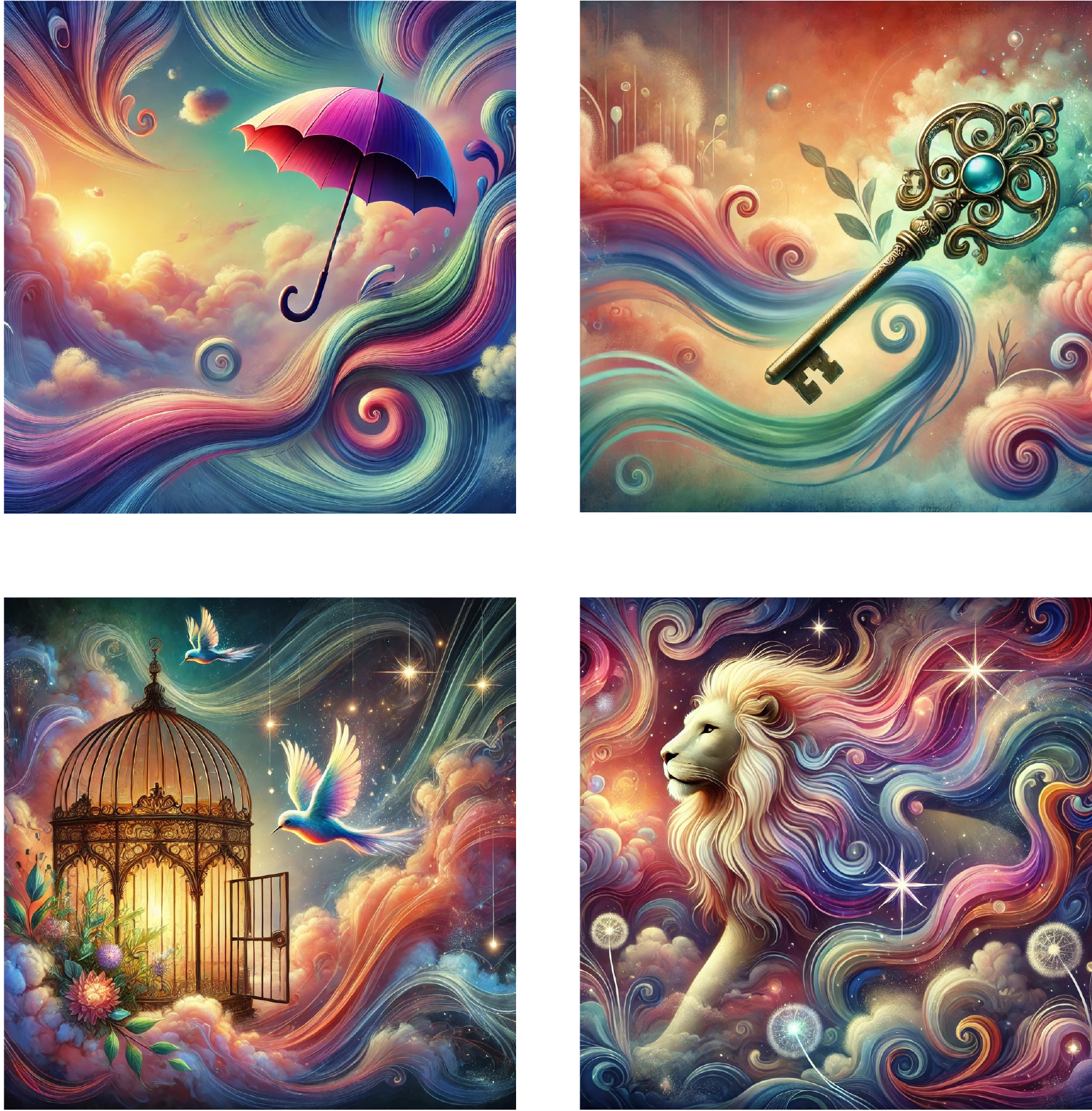}} & Lacking features: surrounded by petals
        \\\\
        \hline
        \\
        \raisebox{-0.05\height}{\includegraphics[width=4.4cm]{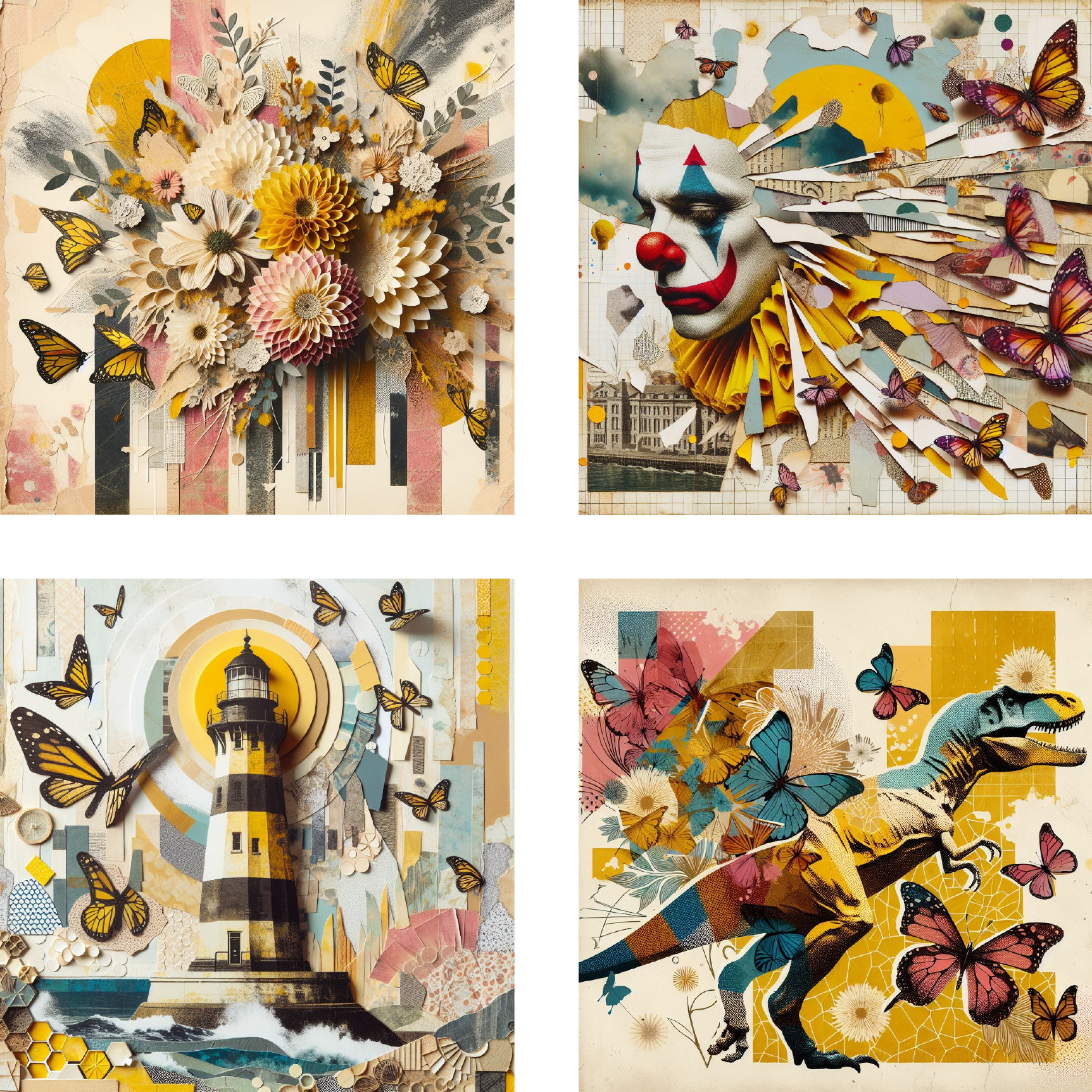}} & 
        \raisebox{-0.05\height}{\includegraphics[width=4.4cm]{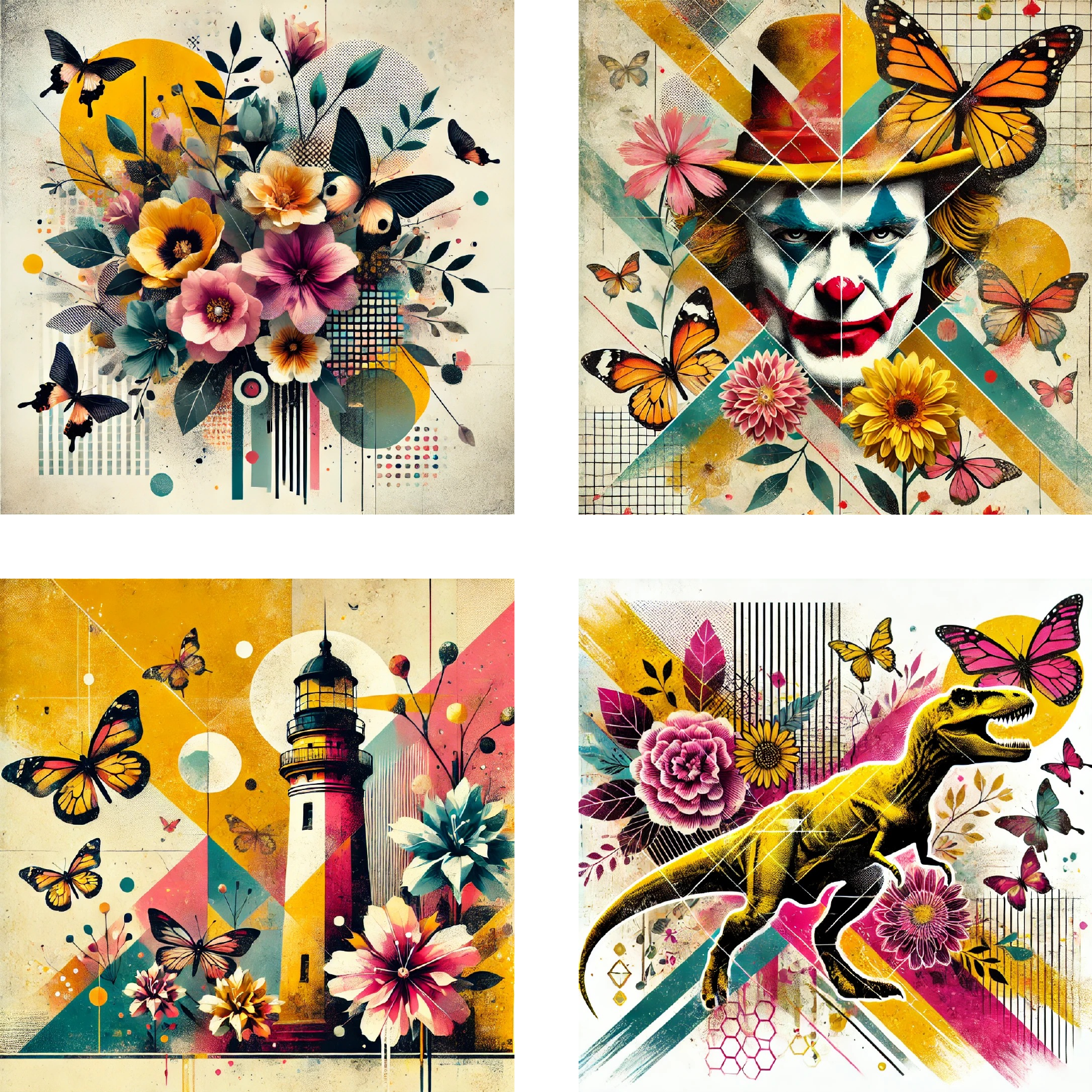}} &
        \raisebox{-0.05\height}{\includegraphics[width=4.4cm]{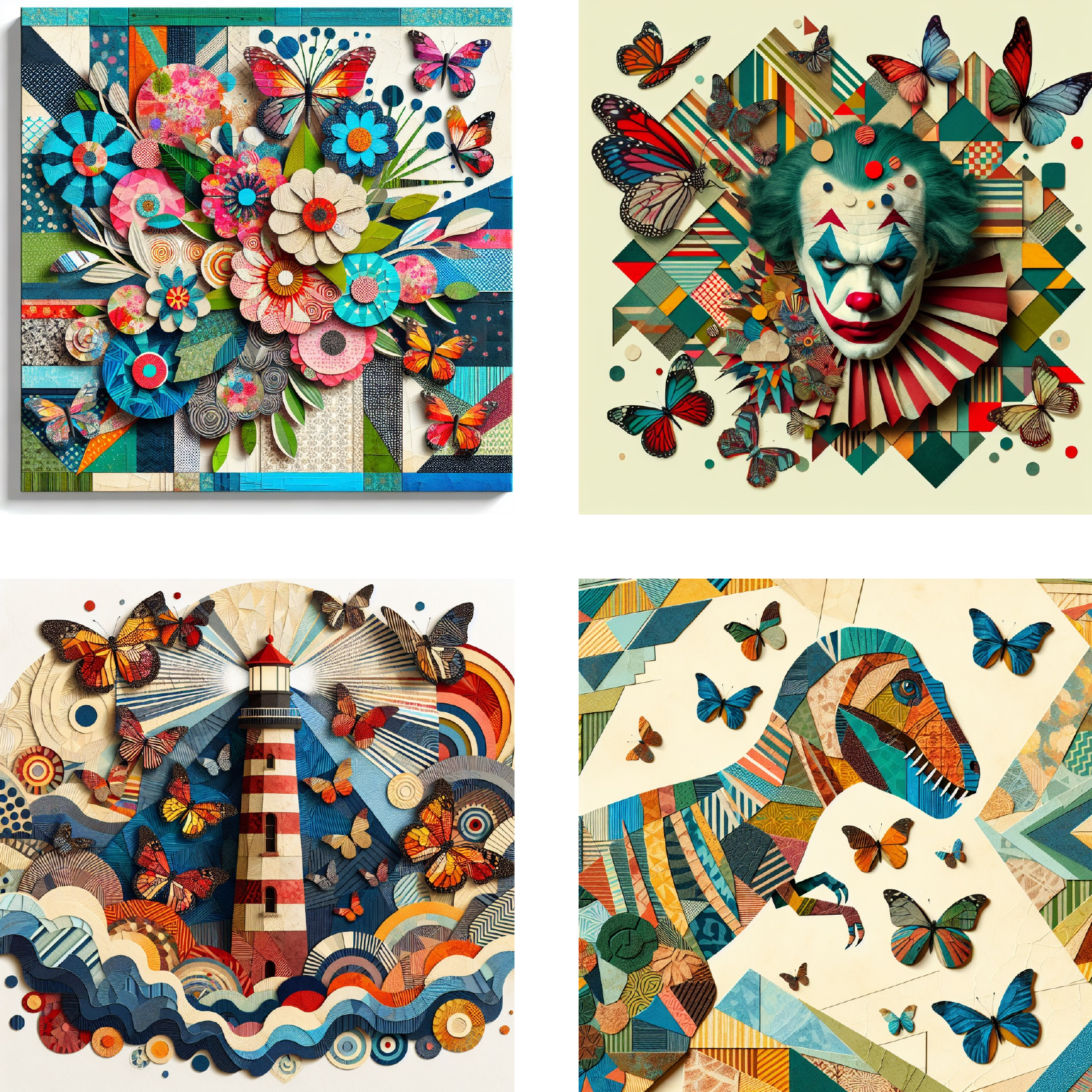}} & Lacking features: dark yellow tone
        \\\\
        \hline
        \\
        \raisebox{-0.05\height}{\includegraphics[width=4.4cm]{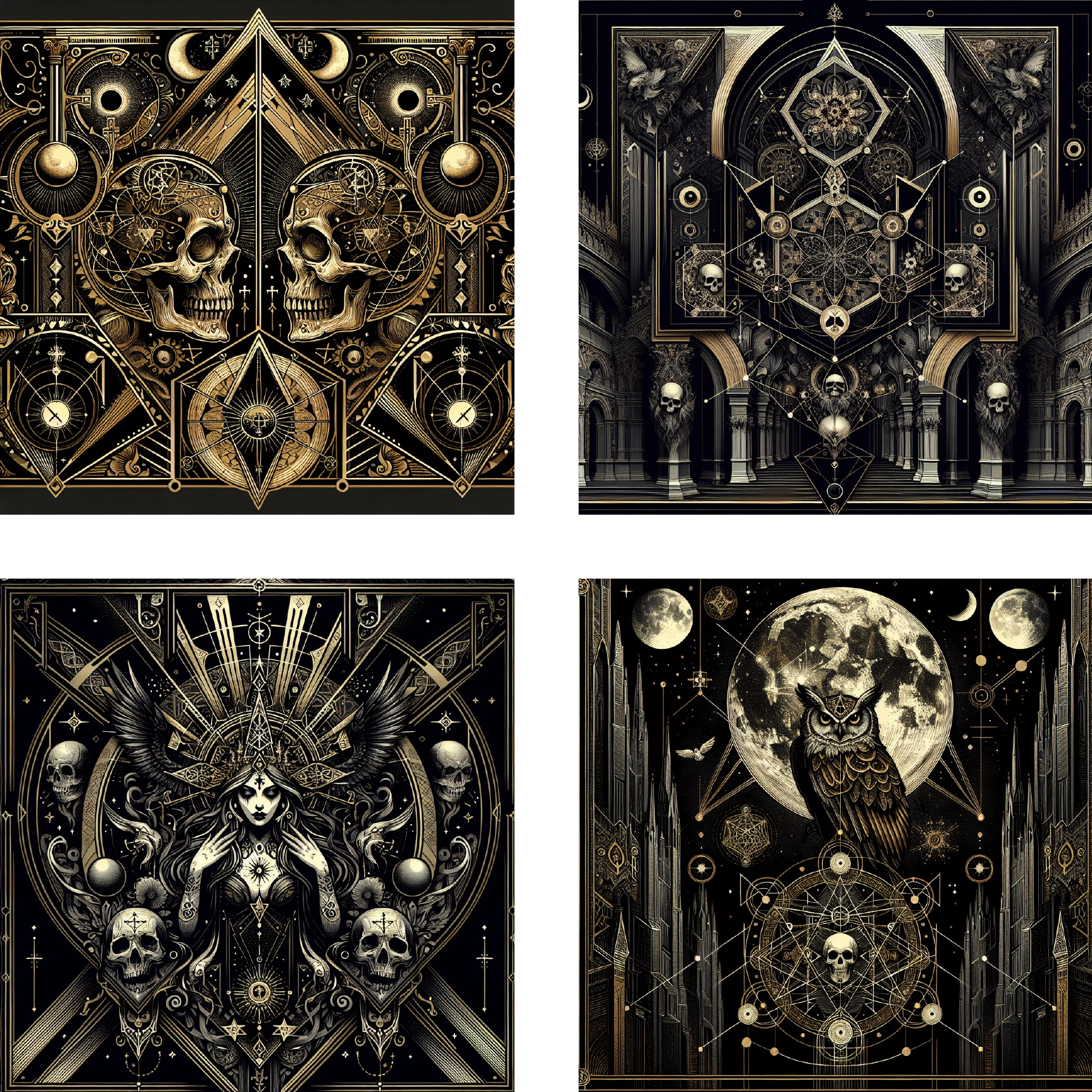}} & 
        \raisebox{-0.05\height}{\includegraphics[width=4.4cm]{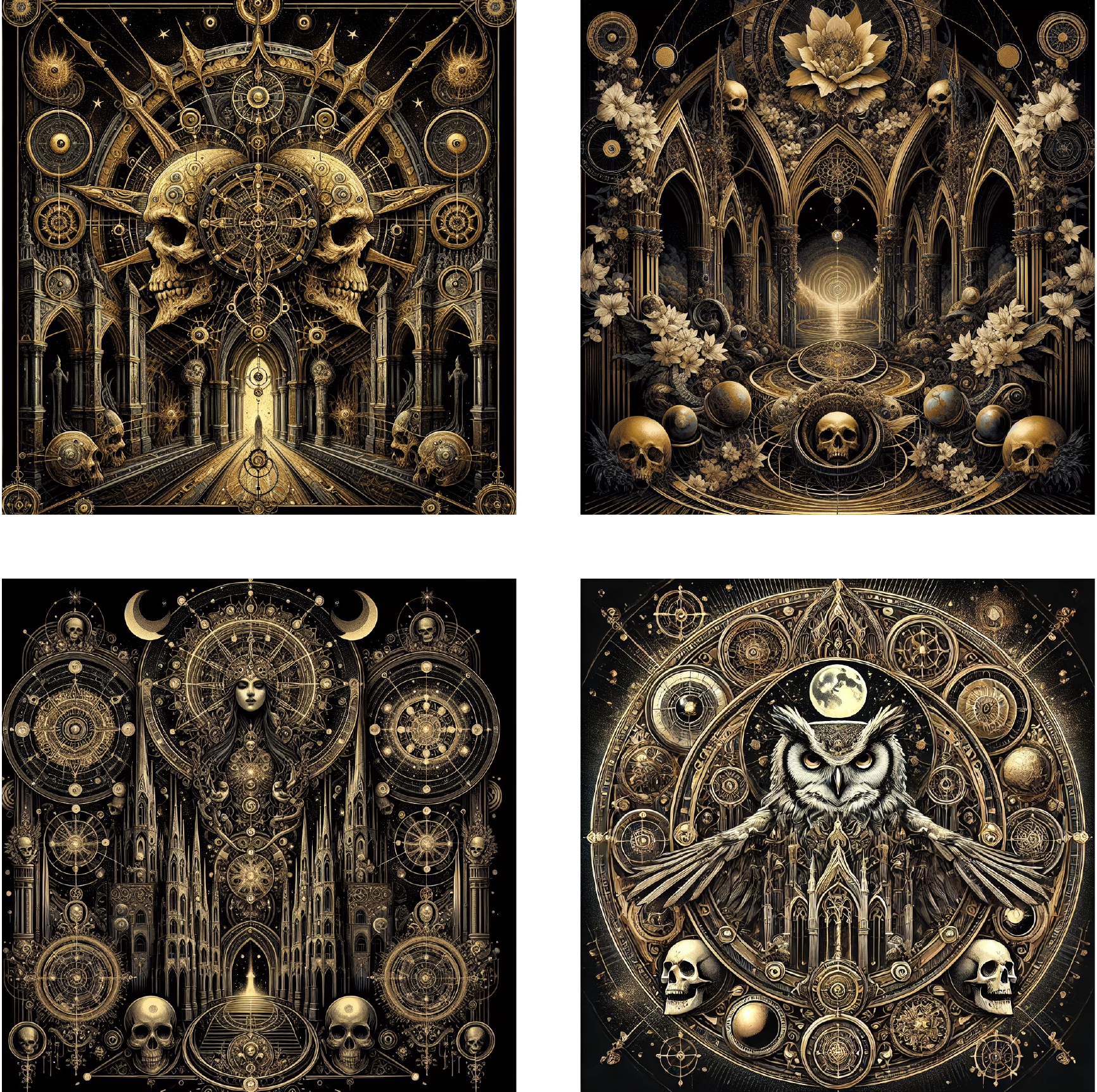}} &
        \raisebox{-0.05\height}{\includegraphics[width=4.4cm]{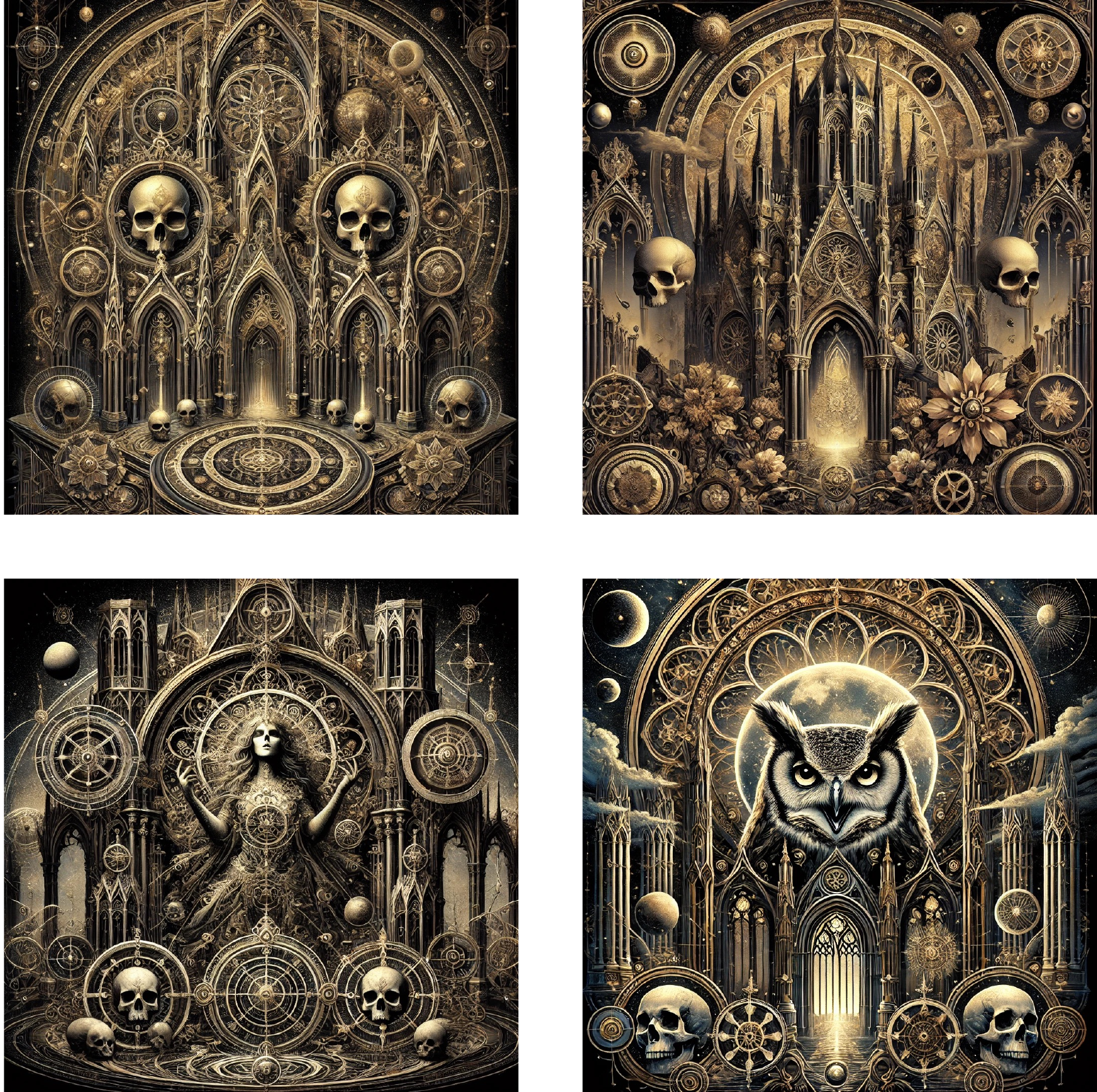}} & Lacking features: symmetry
        \\\\
        \hline
    \end{tabular}
    \end{adjustbox}
    \caption{Three examples are used to demonstrate the impact of removing supplements. "w/o. supp." represents the removal of supplements extracted from the pattern.}
    \label{fig:ablation_comparison}
    
\end{figure*}
Our extraction template incorporates controllable Subjects and Modifiers, complemented by a flexible Supplements module designed to address potential gaps in subject and modifier extraction. Figure~\ref{fig:ablation_comparison} demonstrates the impact of the Supplements module on EvoStealer's effectiveness.

The first case study illustrates how the Supplements module enhances feature detection. While analyzing images individually may cause oversight of shared characteristics—such as the presence of petals in 'a floating umbrella covered in flowers'—the Supplements module successfully captures these overlooked elements in Subject, thereby improving extraction accuracy. In the second case, the module demonstrates its ability to detect visual attributes that are overlooked by predefined modifier categories, such as 'dark yellow tone' within the 'Visual Composition and Structure' categories. The third case exemplifies the module's capacity to identify fundamental aesthetic properties like symmetry, which fall outside established modifier categories. These examples highlight how the Supplements module's flexibility enables the detection of additional key features, ultimately enhancing the quality of image generation.

\section{Evolution Progress} \label{app_progress}
\begin{figure*}[h!]
    \centering
    \includegraphics[width=\linewidth]{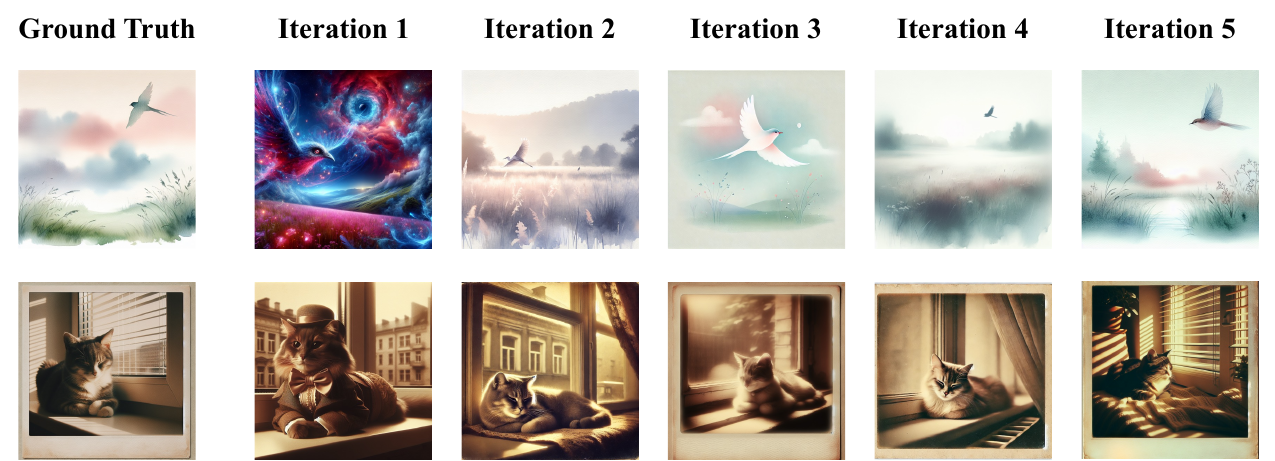}
    \caption{Results of each evolutionary cycle.}
    \label{fig:process}
\end{figure*}
\newcolumntype{C}[1]{>{\centering\arraybackslash}m{#1}}

\begin{figure*}[!ht]
    \vspace{-10pt}
    \centering
    \begin{adjustbox}{width=\textwidth,center}
    \begin{tabular}{C{5cm}C{2cm}C{4cm}C{4cm}}
        \hline
        \textbf{Original Template} & \textbf{Subject} & \textbf{Original Images} & \textbf{EvoStealer} \\
        \hline
        \\
        "Act as you are Arkhip Kuindzhi, and you are living during Industrial Revolution time, you have access only to drawing colors and tools available that time. Create image of [subject]." &
        "cat", "train" &
        \raisebox{-0.05\height}{\includegraphics[width=4cm]{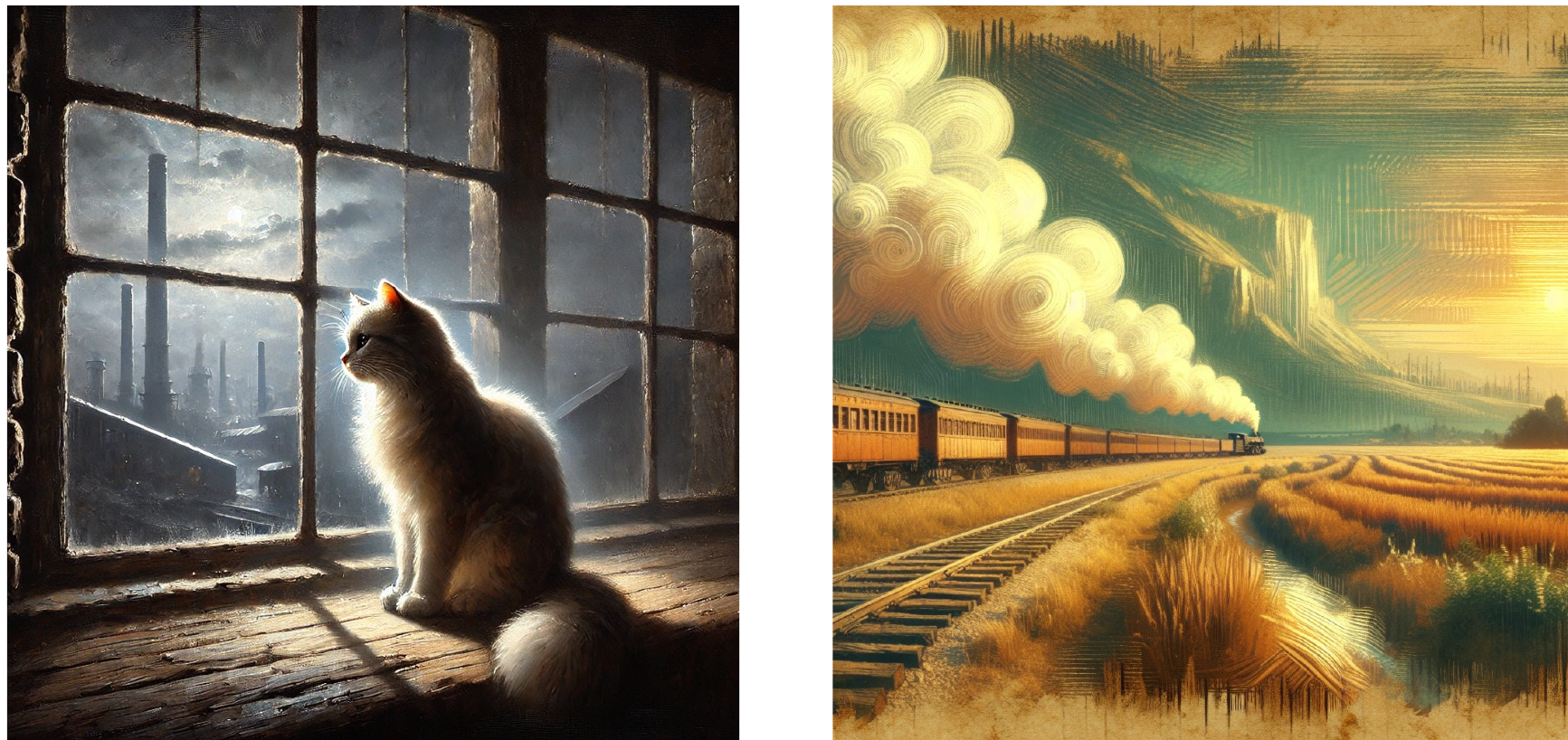}} & 
        \raisebox{-0.05\height}{\includegraphics[width=4cm]{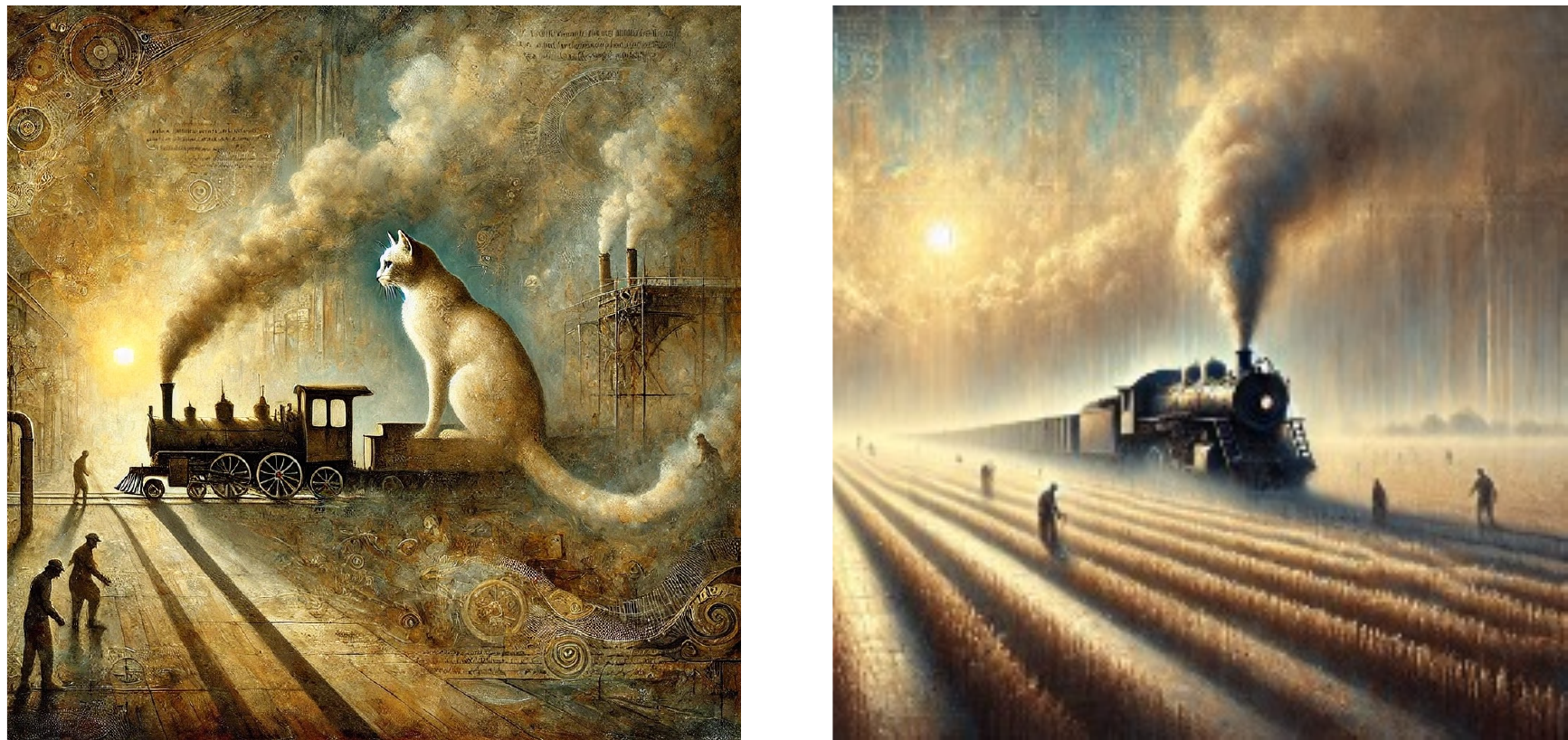}}
        \\\\
        \hline
        \\
        "[subject], logo, funny children's hand drawn style, doodles, minimalism, cute character, in pastel colors including orange and bright blue on a clean black background, hand writing 'hello' with a bold character underneath."&
        "bear", "alligator"
        &
        \raisebox{-0.05\height}{\includegraphics[width=4cm]{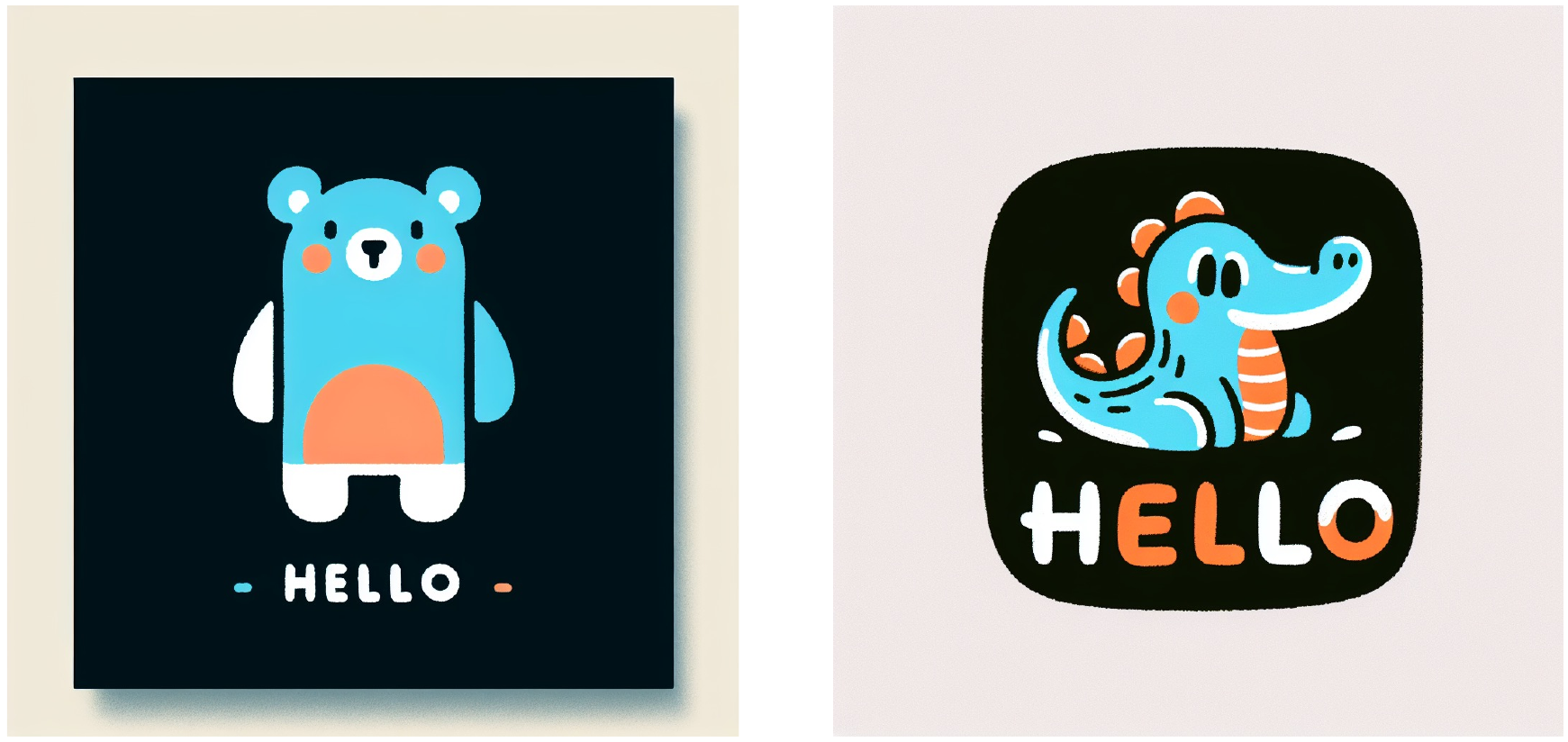}} & 
        \raisebox{-0.05\height}{\includegraphics[width=4cm]{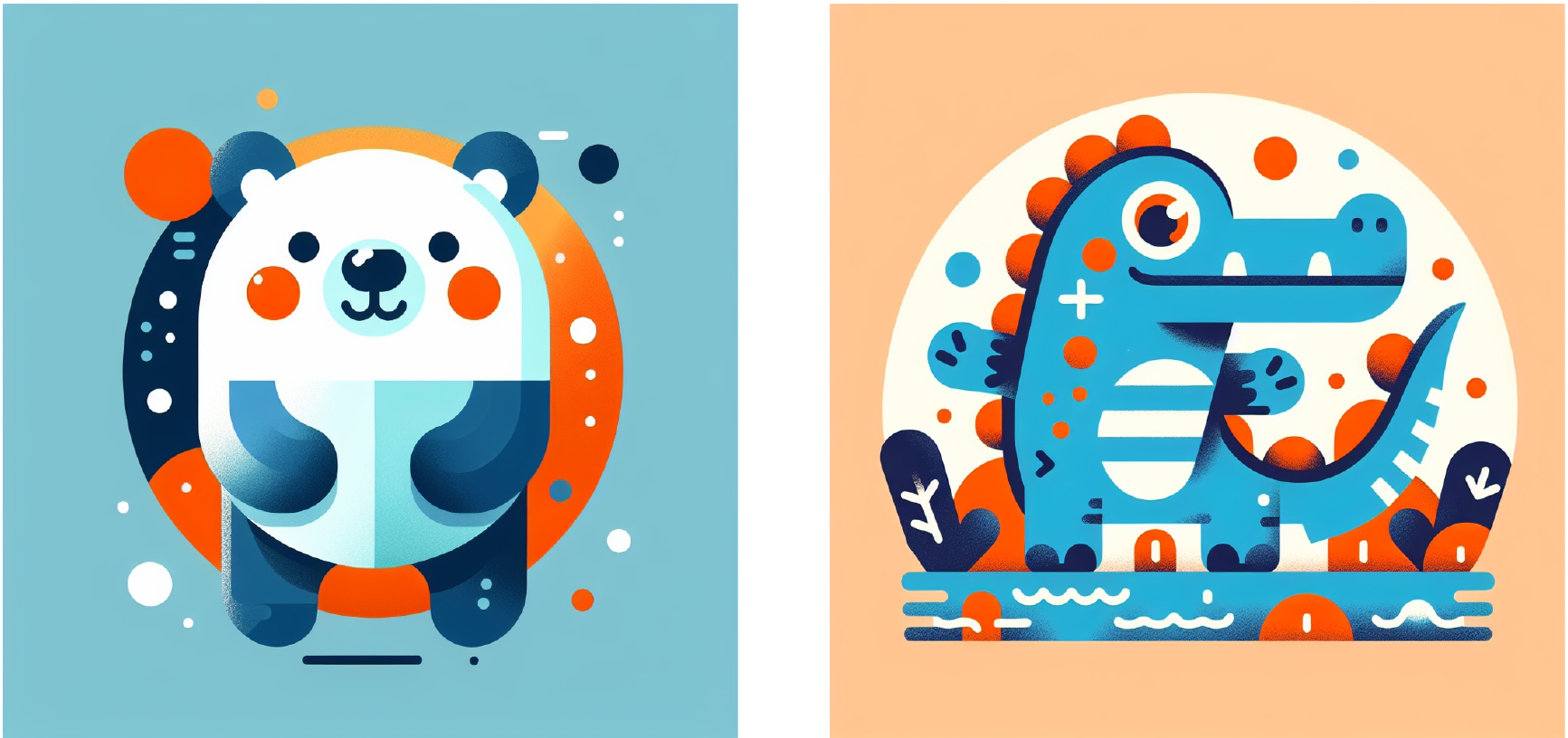}} 
        \\\\
        \hline
        \\
        "[subject] / Dreamlike Ethereal Illustrations / Watercolor-Like Techniques / Loose Expressive Brushstrokes / Cool Pastel Shades / Floating Crystalline Structures / Organic Surreal Shapes / Delicate Flowing Line Work / Tranquil Imaginary Worlds" &
        "Two Heads Are Better Than One", "Moonlit Owl" &
        \raisebox{-0.05\height}{\includegraphics[width=4cm]{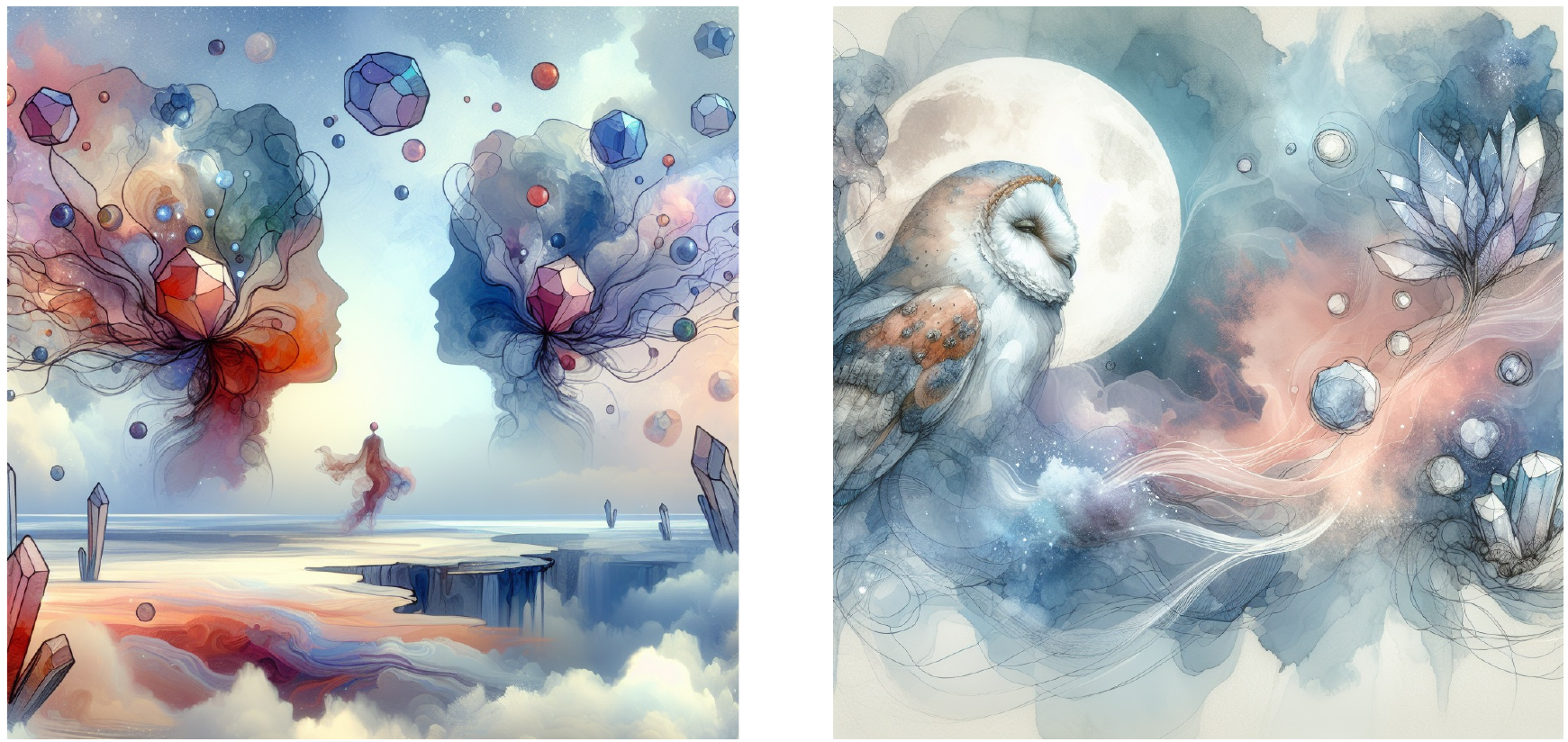}} & 
        \raisebox{-0.05\height}{\includegraphics[width=4cm]{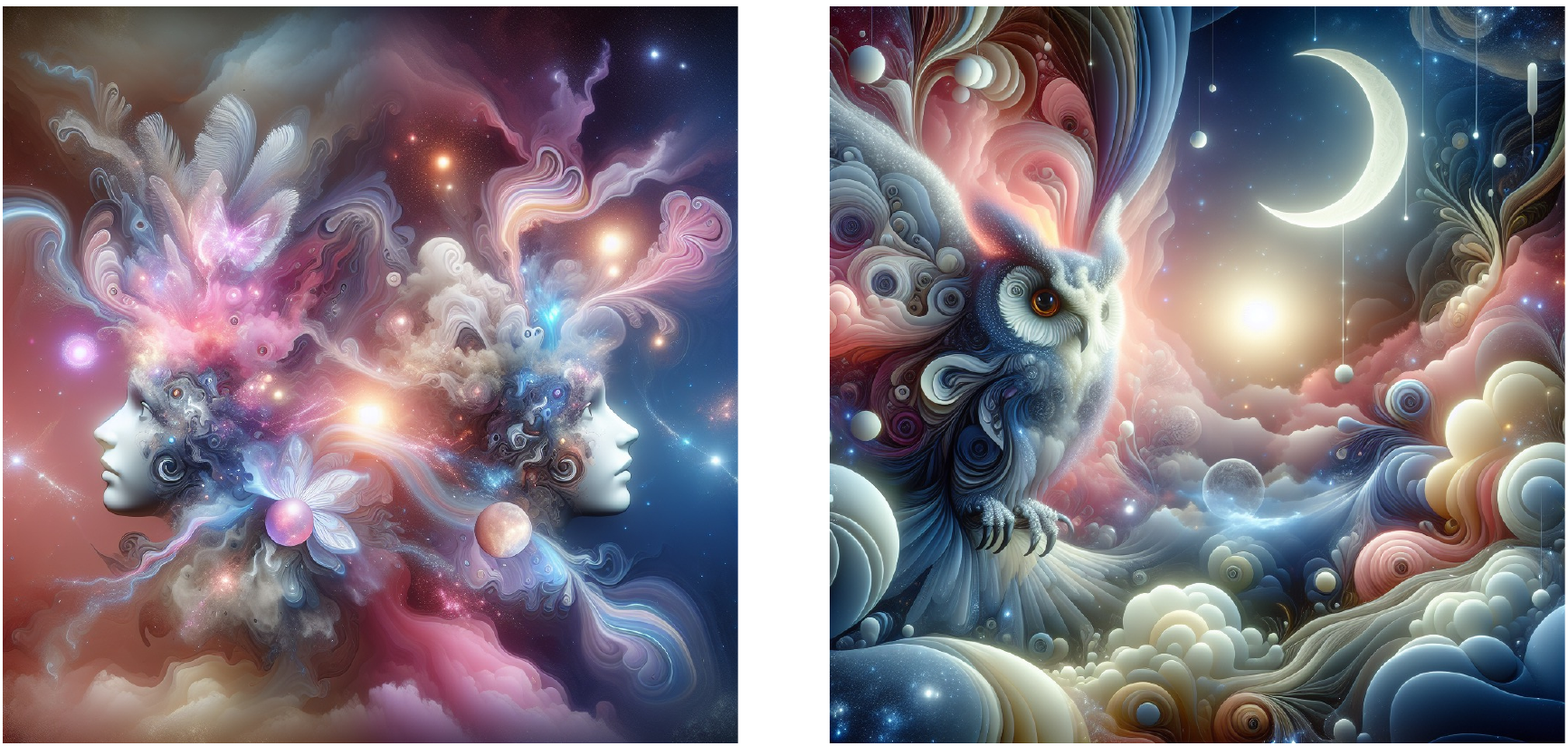}} 
        \\\\
        \hline
    \end{tabular}
    \end{adjustbox}
    \caption{Three failure cases in EvoStealer.}
    \label{fig:app_fail}
    
\end{figure*}

Figure~\ref{fig:process} presents the iterative results of EvoStealer on in-domain data across two distinct styles. The figure demonstrates that with each iteration, the generated images progressively converge toward the ground truth style. This progression indicates that EvoStealer successfully refines the quality of style descriptors throughout its iterative process, resulting in images that increasingly approximate the target style. The visual comparison clearly illustrates the algorithm's capacity to incrementally improve stylistic fidelity through successive refinements.

\section{Cost Estimate} \label{app_cost}
The execution process of EvoStealer comprises three main stages: population initialization, differential evolution (including the fitness function), and image synthesis. We assess the cost from three perspectives: API call frequency, token consumption, and image generation. While API calls and image generation can be accurately and directly measured, token consumption is estimated. Given the instability of the model's output, only the input portion is estimated. For this analysis, we evaluate the cost of stealing a prompt template using EvoStealer, based on GPT-4o.

During the population initialization phase, EvoStealer performs two key operations: image element extraction (which generates <subject, modifiers, supplements> triples) and initial template synthesis. On average, this requires 10 calls to GPT-4o, consuming 1.6k tokens, with an estimated cost of approximately \$0.04. In the differential evolution phase, EvoStealer performs operations such as difference and commonality identification, mutation, mutation addition, and crossover. Additionally, for each offspring, template synthesis and image generation are required for both creation and evaluation. On average, this phase involves 125 API calls, consumes 117.5k tokens, and generates 25 images, resulting in a total cost of approximately \$1.30. In the image synthesis phase, only the optimal template is used to generate 9 images. This requires 9 API calls and 9 image generations, totaling \$0.36. Thus, the overall cost amounts to approximately \$1.70.

\section{Failure Cases} \label{app_failcases}

In this section, we will examine several typical failure cases. These failures stem either from the complexity of the images themselves and vague descriptions, or from the inherent limitations of the current EvoStealer method. Figure~\ref{fig:app_fail} illustrates representative examples.

A primary limitation is the system's inadequate interpretation of specific artistic styles. Analysis of PromptBase and LaPrompt platforms reveals that many prompt templates incorporate stylistic modifiers, such as "Arshile Gorky style," "Disney style," and "Renaissance style." However, the system struggles to accurately identify and replicate the distinctive characteristics of individual artists' techniques or historical artistic movements, resulting in significant stylistic disparities between generated and source images.

A second limitation concerns text recognition capabilities. The current EvoStealer implementation lacks explicit protocols for extracting textual elements from images. Despite MLLMs' inherent text recognition capabilities, this functionality remains underutilized in the present version—a limitation scheduled for address in future iterations.

The third limitation involves comprehensive detail preservation. When processing images with complex color palettes and rich content, EvoStealer may fail to capture fine-grained features, leading to degraded quality in the resultant prompt templates.

\end{document}